\documentclass[10pt,twocolumn,letterpaper]{article}

\usepackage{cvpr}
\usepackage{times}
\usepackage{epsfig}
\usepackage{graphicx}
\usepackage{amsmath}
\usepackage{amssymb}

\usepackage{subfigure}
\usepackage[breaklinks=true,bookmarks=false]{hyperref}

\cvprfinalcopy % *** Uncomment this line for the final submission

\def\cvprPaperID{****} % *** Enter the CVPR Paper ID here

\ifcvprfinal\fi
\begin{document}

\newcommand{\PutCap}{{{~(a)~~~~~~~~~~~~~~~~~~~~~~~~~~~~~~~~~~~~~~~~~~(b)~~~~~~~~~~~~~~~~ }}}

%%%%%%%%% TITLE
\title{ROI Pooled Correlation Filters for Visual Tracking}

\author{Yuxuan Sun$^1$, Chong Sun$^2$, Dong Wang$^1$\thanks{Corresponding Author: Dr. Wang.}, You He$^3$, Huchuan Lu$^{1,4}$\\
$^1$School of Information and Communication Engineering, Dalian University of Technology, China\\
$^2$Tencent Youtu Lab, China\\
$^3$Naval Aviation University, China\\
$^4$Peng Cheng Laboratory, China\\
{\tt\small rumsyx@mail.dlut.edu.cn, waynecsun@tencent.com, heyou\_f@126.com,
\{wdice,lhchuan\}@dlut.edu.cn}
% \and
% Chong Sun\\
}

\maketitle
\thispagestyle{empty}
%%%%%%%%% ABSTRACT
\begin{abstract}
The ROI (region-of-interest) based pooling method performs pooling operations on the cropped ROI regions for various samples and has shown great success in the object detection methods. It compresses the model size while preserving the localization accuracy, thus it is useful in the visual tracking field. Though being effective, the ROI-based pooling operation is not yet considered in the correlation filter formula.
In this paper, we propose a novel ROI pooled correlation filter (RPCF) algorithm for robust visual tracking. Through mathematical derivations, we show that the ROI-based pooling can be equivalently achieved by enforcing additional constraints on the learned filter weights, which makes the ROI-based pooling feasible on the virtual circular samples. Besides, we develop an efficient joint training formula for the proposed correlation filter algorithm, and derive the Fourier solvers for efficient model training. Finally, we evaluate our RPCF tracker on OTB-2013, OTB-2015 and VOT-2017 benchmark datasets. Experimental results show that our tracker performs favourably against other state-of-the-art trackers.

\end{abstract}

%%%%%%%%% BODY TEXT
\section{Introduction}

Visual tracking aims to localize the manually specified target object in the successive frames, and it has been densely studied in
the past decades for its broad applications in the automatic drive, human-machine interaction, behavior recognition, etc. Till now, visual tracking
is still a very challenging task due to the limited training data and plenty of real-world challenges, such as occlusion, deformation and illumination variations.

\begin{figure}
\centering
\includegraphics[width=1\linewidth]{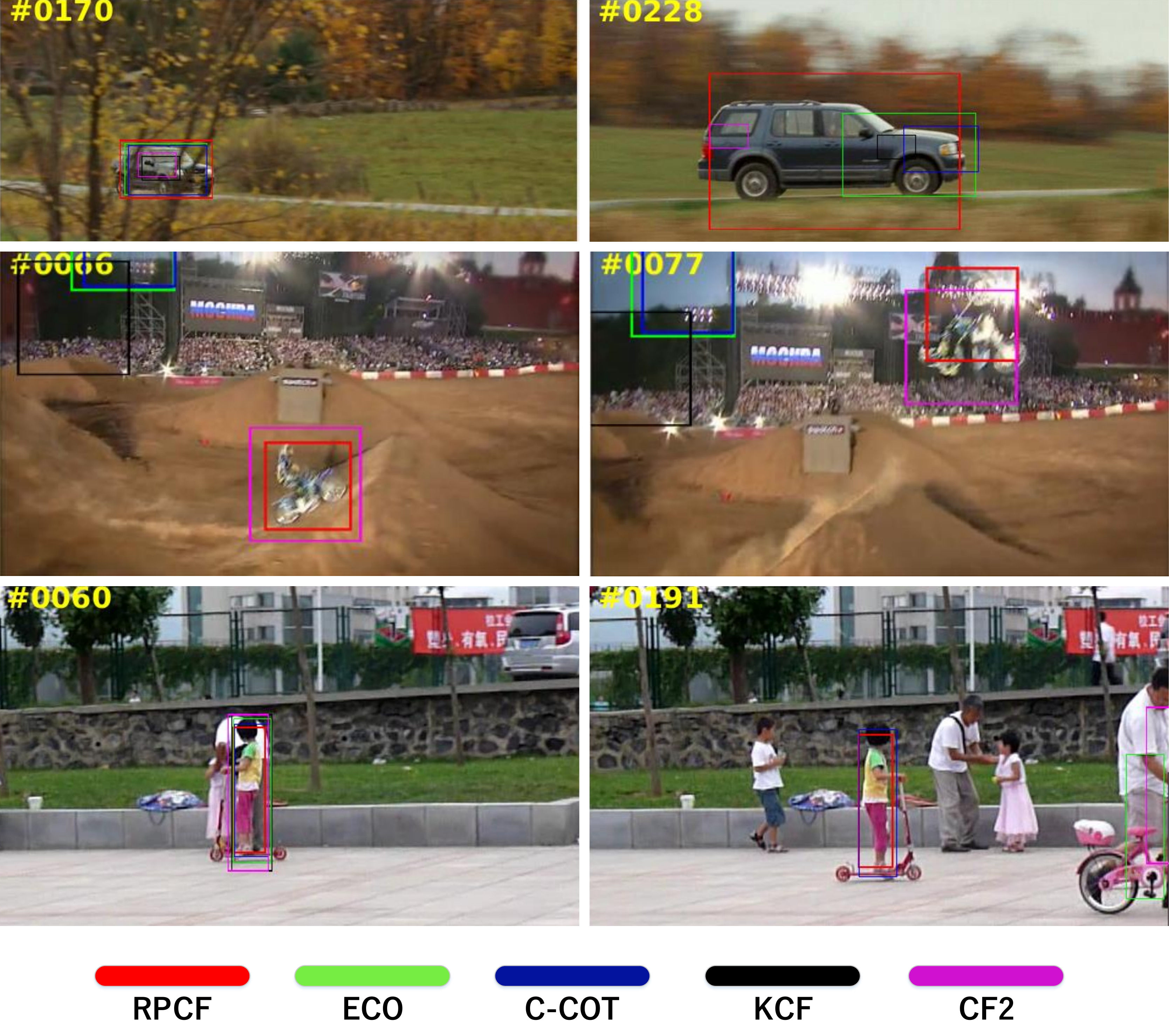}\\
\caption{Visualized tracking results of our method and other four competing algorithms.
Our tracker performs favourably against the state-of-the-art.}
\label{fig:homepage_demo}
\end{figure}

In recent years, the correlation filter (CF) has become one of the most widely used formulas in visual tracking for its computation efficiency. 
The success of the correlation filter mainly comes from two aspects: first, 
by exploiting the property of circulant matrix, the CF-based algorithms do not need to construct the training and testing samples explicitly, and can be 
efficiently optimized in the Fourier domain, enabling it to handle more features;  
second, optimizing a correlation filter can be equivalently converted to solving a system of linear functions, thus the filter weights can 
either be obtained with the analytic solution (e.g.,~\cite{danelljan2015learning,danelljan2014accurate}) or be solved via the optimization algorithms with quadratic convergence~\cite{danelljan2015learning, danelljan2017eco}. As is well recognized, the primal correlation filter algorithms have limited tracking performance due to the boundary effects and the over-fitting problem. The phenomenon of boundary effects is caused by the periodic assumptions of the training samples, while the over-fitting problem
is caused by the unbalance between the numbers of model parameters and the training samples.
Though the boundary effects have been well addressed in several recent papers (\eg, SRDCF~\cite{danelljan2015learning}, DRT~\cite{sun2018correlation}, BACF~\cite{galoogahi2017learning} and ASRCF~\cite{dai2019}), the over-fitting problem is still not paid much attention
to and remains to be a challenging research hotspot. 

The average/max-pooling operation has been widely used in the deep learning methods via the pooling layer, which is shown to be effective in handling the over-fitting problem and deformations.
Currently, two kinds of pooling operations are widely used in deep learning methods. The first one performs average/max-pooling on the entire input feature map and obtains a 
feature map with reduced spatial resolutions.
In the CF formula, the pooling operation on the input feature map can lead to fewer available synthetic training samples, which limits the discriminative ability of the learned filter. Also, the smaller size of the feature map will significantly influence the localization accuracy.
However, the ROI (Region of Interest)-based pooling operation is an alternative, which has been successfully embedded into several object detection networks (\eg, \cite{girshick2015fast,ren2015faster}).
Instead of directly performing the average/max-pooling on the entire feature map, the ROI-based pooling method first crops large numbers of ROI regions, each of which corresponds to a target candidate, and then performs average/max-pooling for each candidate ROI region independently.
The ROI-based pooling operation has the merits of a pooling operation as mentioned above, and at the same time retains the number of training samples and the spatial information for localization, thus it is meaningful to introduce the ROI-based pooling into the CF formula.
Since the CF algorithm has no access to real-world samples, it remains to be investigated on how to exploit the ROI-based pooling in a correlation filter formula.

In this paper, we study the influence of the pooling operation in visual tracking, and propose a novel ROI pooled correlation filters algorithm. Even though the ROI-based pooling algorithm has been successfully applied in many deep learning-based applications, it is seldom considered in the visual tracking field, especially in the correlation filter-based methods. Since the correlation filter formula does not really extract positive and negative samples, it is infeasible to perform the ROI-based pooling like Fast R-CNN~\cite{girshick2015fast}. Through mathematical derivation, we provide an alternative solution to implement the ROI-based pooling.
We propose a correlation filter algorithm with equality constraints,
through which the ROI-based pooling can be equivalently achieved. We propose an Alternating Direction Method Of Multipliers (ADMM) algorithm to solve the optimization problem, and provide an efficient solver in the Fourier domain. Large number of experiments on the OTB-2013~\cite{wu2013online}, OTB-2015~\cite{wu2015object} and VOT-2017~\cite{VOT2017} datasets validate the effectiveness of the proposed method (see Figure~\ref{fig:homepage_demo} and Section~\ref{sec:experiments}).
The contributions of this paper are three-fold:
\begin{itemize}
\item This paper is the first attempt to introduce the idea of ROI-based pooling in the correlation filter formula. It proposes a correlation filter algorithm with equality constraints, through which the ROI-based pooling operation can be equivalently achieved without the need for real-world ROI sample extraction. The learned filter weights are insusceptible to the over-fitting problem and are more robust to deformations.

\item This paper proposes a robust ADMM method to optimize the proposed correlation filter formula in the Fourier domain. With the computed Lagrangian multipliers, the paper aims to use the conjugate gradient method for filter learning, and develops efficient optimization strategy for each step.

\item This paper conducts large amounts of experiments on three available public datasets. The experimental results validate the effectiveness of the proposed method. Project page :  https://github.com/rumsyx/RPCF.

\end{itemize}

%-------------------------------------------------------------------------
\section{Related Work}
The recent papers on visual tracking are mainly based on the correlation filters and deep networks~\cite{li2018deep}, many of which have impressive performance. In this section, we primarily focus on the algorithms based on the correlation filters and briefly introduce related issues of the pooling operations.

\textbf{Discriminative Correlation Filters.} 
Trackers based on correlation filters have been the focus of researchers in recent years, which have achieved the top performance in various datasets. The correlation filter algorithm in visual tracking can be dated back to the MOSSE tracker~\cite{Bolme2010Visual}, which takes the single-channel gray-scale image as input. Even though the tracking speed is impressive, the accuracy is not satisfactory. Based on the MOSSE tracker, Henriques~\etal advance the state-of-the-art by introducing the kernel functions~\cite{henriques2012exploiting} and higher dimensional features~\cite{henriques2015high}.
Ma~\etal~\cite{ma2015hierarchical} exploit the rich representation information of deep features in the correlation filter formula, and fuse the responses of various convolutional features via a coarse-to-fine searching strategy.
Qi~\etal~\cite{qi2016hedged} extend the work of~\cite{ma2015hierarchical} by exploiting the Hedge method to learn the importance for each kind of feature adaptively. Apart from the MOSSE tracker, the aforementioned algorithms learn the filter weights in the dual space, which have been attested to be less effective than the primal space-based algorithms~\cite{danelljan2014accurate,danelljan2015learning,henriques2015high}. However, correlation filters learned in the primal space are severely influenced by the boundary effects and the over-fitting problem. Because of this, Danelljan~\etal~\cite{danelljan2015learning} introduce a weighted regularization constraint on the learned filter weights, encouraging the algorithm to learn more weights on the central region of the target object. The SRDCF tracker~\cite{danelljan2015learning} has become a baseline algorithm for many latter trackers, \eg, CCOT~\cite{danelljan2016beyond} and SRDCFDecon~\cite{danelljan2016adaptive}. The BACF tracker~\cite{galoogahi2017learning} provides another feasible way to address the boundary effects, which generates real-world training samples and greatly improves the discriminant power of the learned filter. 
Though the above methods have well addressed the boundary effects, the over-fitting problem is rarely considered. The ECO tracker~\cite{danelljan2017eco} jointly learns a projection matrix and the filter weights, through which the model size is greatly compressed. Different from the ECO tracker, our method introduces the ROI-based pooling operation into a correlation filter formula, which does not only address the over-fitting problem but also makes the learned filter weights more robust to deformations.

\textbf{Pooling Operations.} The idea of the pooling operation has been used in various fields in computer vision, \eg, feature extraction~\cite{dalal2005histograms,lowe2004distinctive}, convolutional neural networks~\cite{simonyan2014very,he2016deep}, to name a few. 
Most of the pooling operations are performed on the entire feature map to either obtain more stable feature representations or rapidly compress the model size.  In~\cite{dalal2005histograms}, Dalal~\etal
divide the image window into dozens of cells, and compute the histogram of gradient directions in each divided cell. The computed feature representations are more robust than the ones based on individual pixels. 
In most deep learning-based algorithms (\eg,~\cite{dalal2005histograms,lowe2004distinctive}), the pooling operations are performed via a pooling layer, which accumulates the multiple response activations over a small neighbourhood region. The localization accuracy of the network usually decreases after the pooling operation. Instead of the primal max/average-pooling layer, the faster R-CNN method~\cite{girshick2015fast} exploits the ROI pooling layer to ensure the localization accuracy and at the same time compress the model size. The method firstly extracts the ROI region for each candidate target object via a region of proposal network (RPN), and then performs the max-pooling operation on the ROI region to obtain more robust feature representations. Our method is inspired by the ROI pooling proposed in~\cite{girshick2015fast}, and is the first attempt to introduce the ROI-based pooling operation into the correlation filter formula.

%-------------------------------------------------------------------------

\section{Correlation Filter and Pooling}
In this section, we briefly revisit the two key technologies closely related to our approach (\ie, the correlation filter and pooling operation).

\subsection{Revisit of Correlation Filter }
\label{sec:bacf}
To help better understand our method, we first introduce the primal correlation filter algorithm. Given an input feature map, a correlation filter algorithm aims at learning a set of filter weights to regress the Gaussian-shaped response.
We use $y_d\in \mathbb{R}^{N}$ to denote the desired Gaussian-shaped response, and $x$ to denote the input feature map with $D$ feature channels $x_1, x_2,...,x_D$. For each feature channel $x_d\in \mathbb{R}^{N}$, a correlation filter algorithm computes the response by convolving $x_d$ with the filter weight $w_d\in\mathbb{R}^{N} $. Based on the above-mentioned definitions and descriptions,
the optimal filter weights can be obtained by optimizing the following objective function:
\begin{equation}
    E(w) =  \frac{1}{2}\left\| y - \sum\limits_{d = 1}^{D} {w_d*x_d } \right\|_2^2  +  \frac{\lambda}{2} \sum\limits_{d = 1}^{{D}} \left\| {{w_d}} \right\|_2^2,
    \label{equ:primal_cf}
\end{equation}
where $*$ denotes the circular convolution operator,
$w = [w_1, w_2, ..., w_D]$ is concatenated filter vector, $\lambda$ is a trade-off parameter to balance the importance between the regression and the regularization losses. According to the Parseval's theorem,
Eq.~\ref{equ:primal_cf} can be equivalently written in the Fourier domain as
\begin{equation}
    E(\hat w) =  \frac{1}{2}\left\| {\hat y - \sum\limits_{d = 1}^D {{{\hat w}_d}\odot{{\hat x}_d}} } \right\|_2^2 +  \frac{\lambda}{2}\sum\limits_{d = 1}^{{D}}\left\| {{{\hat w}_d}} \right\|_2^2,
\end{equation}
where  $ \odot $ is the Hadamard product.
We use $\hat y$, $\hat w_d$, $\hat x_d$ to denote the Fourier domain of vector $y$, $w_d$ and $x_d$.

\begin{figure}[t]
    \centering
    \includegraphics[width=1\linewidth]{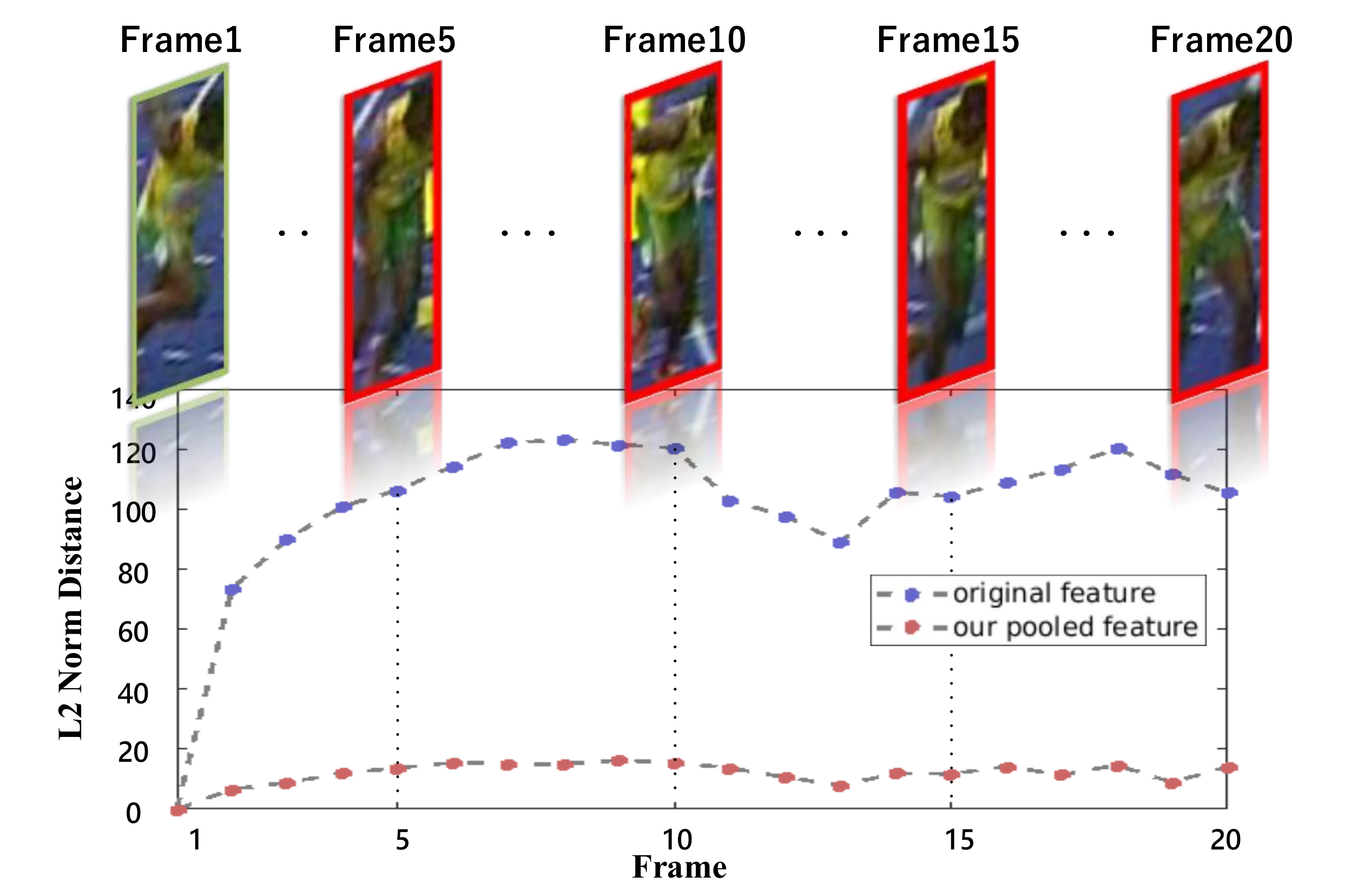}\\
    \caption{Illustration showing that ROI pooled features are more robust to target deformations than the original ones. For both features, we compute the $\ell_2$ loss
    between features extracted from Frames 2-20 and Frame 1, and visualize the distances via red and blue dots respectively. }
    \label{fig:pooling_robust}
\end{figure}

\subsection{Pooling Operation in Visual Tracking}
\begin{figure}[t]
  \centering
  \includegraphics[width=1\linewidth]{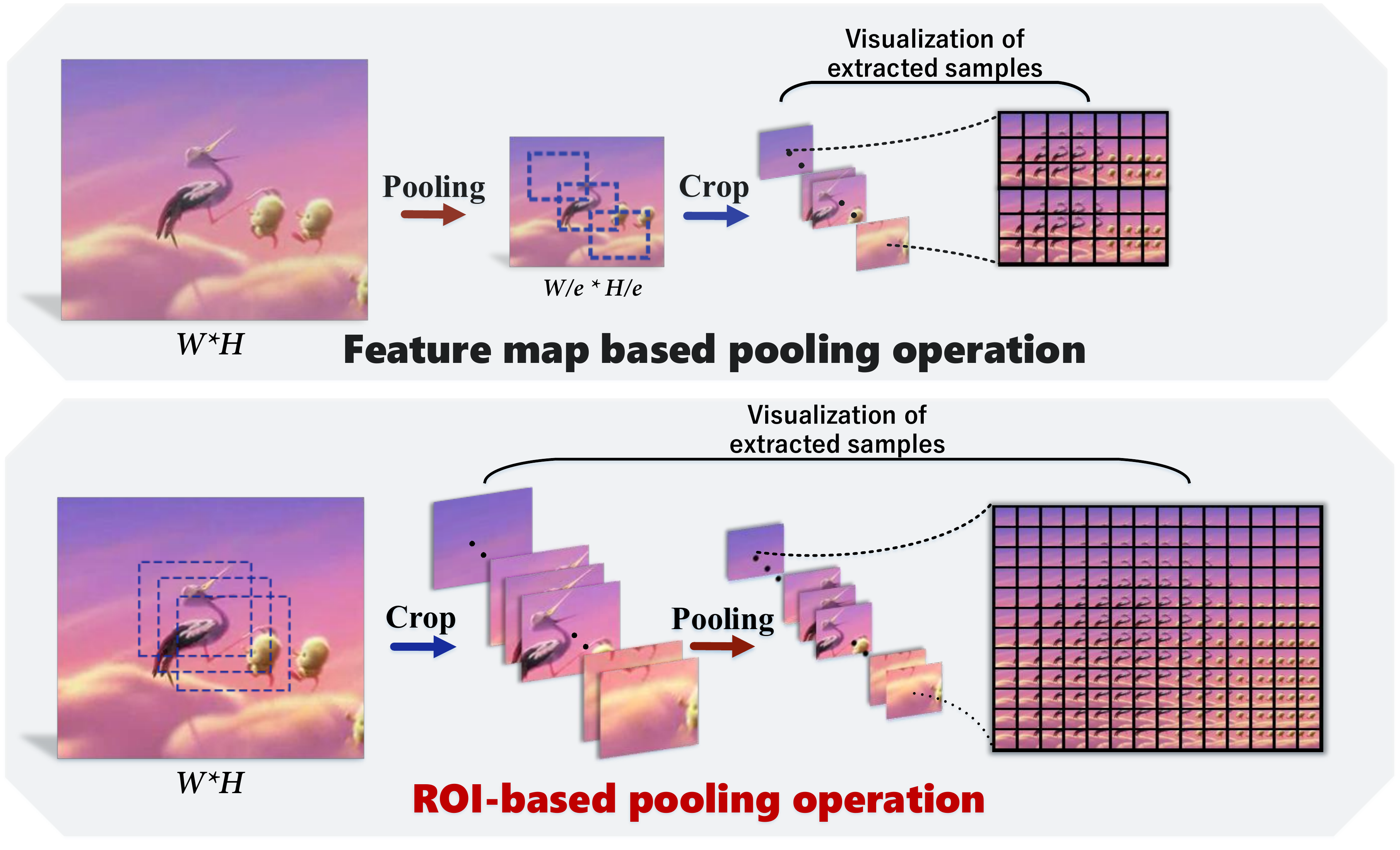}\\
  \caption{Illustration showing the difference between the feature map based and the ROI-based pooling operations. 
  For clarity, we use 8 as the stride for sample extraction on the original image. This corresponds to a stride = 2 feature extraction in the HOG feature with 4 as the cell size. The pooling kernel size is set as e = 2 in this example.
  }
  \label{fig:pooling_demo}
\end{figure}

As is described by many deep learning methods~\cite{simonyan2014very,gatys2016image}, the pooling layer plays a crucial rule in addressing the over-fitting problem. 
Generally speaking, a pooling operation tries to fuse the neighbourhood response activations into one, through which the model parameters can be effectively compressed. In addition to addressing the over-fitting problem, the pooled feature map becomes more robust to deformations (Figure~\ref{fig:pooling_robust}).
Currently, two kinds of pooling operations are widely used, \ie, the pooling operation based on the entire feature map (\eg,~\cite{simonyan2014very,he2016deep}) and the pooling operation based on the candidate ROI region (\eg~\cite{ren2015faster}). The former one has been widely used in the CF trackers with deep features, as a contrast, the ROI-based pooling operation is seldom considered. As is described in Section 1, directly performing average/max-pooling on the input feature map will result in fewer training/testing samples and worse localization accuracy.
We use an example to show how different pooling methods influence the sample extraction process in Figure~\ref{fig:pooling_demo}, wherein 
the extracted samples are visualized on the right-hand side. For simplicity, this example is based on the dense sampling process. The conclusion is also applicable to the correlation filter method, which is essentially trained via densely sampled circular candidates.
In the feature map based pooling operation, the feature map size is first reduced to $W/e\times H/e$, thus leading to fewer samples. However, the ROI-based pooling first crop samples from the $W \times H$ feature map and then performs pooling operations upon them, thus does not influence the training number.
Fewer training samples will lead to inferior discrimination ability of the learned filter, while fewer testing samples will result in inaccurate target localizations. Thus, it is meaningful to introduce the ROI-based pooling operation into the correlation filter algorithms. Since the max-pooling operation will introduce the non-linearity that makes the model intractable to be optimized, the ROI-based average-pooling operation is preferred in this paper.

%-------------------------------------------------------------------------
\section{Our Approach}
\subsection{ROI Pooled Correlation Filter}
In this section, we propose a novel correlation tracking method with ROI-based pooling operation.
Like the previous methods~\cite{henriques2012exploiting,danelljan2016beyond}, we introduce our CF-based tracking algorithm in the one-dimensional domain, and the conclusions can be easily generalized to higher dimensions.
Since the correlation filter does not explicitly extract the training samples, it is impossible to perform the ROI-based pooling operation following the pipeline in Figure~\ref{fig:pooling_demo}.
In this paper, we derive that the ROI-based pooling operation can be implemented by adding additional constraints on the learned filter weights. 

Given a candidate feature vector $v$ corresponding to the target region with $L$ elements,
we perform the average-pooling operation on it with the pooling kernel size $e$. For simplicity, we set $L=eM$, where $M$ is a positive integer (the padding operation can be used if $L$ cannot be divided by $e$ evenly).
The pooled feature vector ${v'}\in\mathbb{R}^M$ can be computed as $v' = \frac{1}{e}Uv$, where the matrix $U\in \mathbb{R}^{M\times Me}$ is constructed as:
\begin{equation}
U = \left[ {\begin{array}{*{20}{c}}
{{{\bf{1}}^e}}&{{{\bf{0}}^e}}& \cdots &{{{\bf{0}}^e}}&{{{\bf{0}}^e}}\\
{{{\bf{0}}^e}}&{{{\bf{1}}^e}}& \cdots &{{{\bf{0}}^e}}&{{{\bf{0}}^e}}\\
 \vdots & \vdots & \ddots &{{{\bf{0}}^e}}&{{{\bf{0}}^e}}\\
{{{\bf{0}}^e}}&{{{\bf{0}}^e}}& \cdots &{{{\bf{1}}^e}}&{{{\bf{0}}^e}}\\
{{{\bf{0}}^e}}&{{{\bf{0}}^e}}& \cdots &{{{\bf{0}}^e}}&{{{\bf{1}}^e}}
\end{array}} \right],
\end{equation}
where ${\bf{1}}^e\in \mathbb{R}^{1\times e}$ denotes a vector with all the entries set as 1, and 
${\bf{0}}^e\in \mathbb{R}^{1\times e}$ is a zero vector. Based on the pooled vector, we compute the response as:
\begin{equation}
    r = {w'^\top}v' = {w'^\top}Uv/e = {\left( {{U^\top}w'} \right)^\top}v/e,
\end{equation}
wherein $w'$ is the weight corresponding to the pooled feature vector, ${U^\top}w' = {[w'(1){{\bf{1}}^e},w'(2){{\bf{1}}^e},...,w'(M){{\bf{1}}^e}]^\top}$. It is easy to conclude that 
average-pooling operation can be equivalently achieved by 
constraining the filter weights in each pooling kernel to have the same value. Based on the discussions above, we define our ROI pooled correlation filter as follows:
\begin{equation}
\begin{array}{l}
{\hspace{-2mm}}
E(w) =  \frac{1}{2}\left\| {y - \sum\limits_{d = 1}^D {({p_d}\odot {w_d})*{x_d}} } \right\|_2^2 + \frac{\lambda}{2}\sum\limits_{d = 1}^D {\left\| {{g_d} \odot {w_d}} \right\|_2^2} \\
{\kern 20pt}{\rm{s}}{\rm{.t}}{\rm{.}}{\kern 1pt} {\kern 1pt} {\kern 1pt} {\kern 1pt} {\kern 1pt} {\kern 1pt} {w_d}(i_\eta) = {w_d}(j_\eta),{\kern 1pt} {\kern 1pt} {\kern 1pt} {\kern 1pt} {\kern 1pt} {\kern 1pt} (i_\eta,j_\eta) \in  \mathcal{P}, \eta = 1,..,K 
\end{array}
\label{eq:optimization_primal}
\end{equation}
where we consider $K$ equality constraints to ensure that filter weights in each pooling kernel have the same value, $\mathcal{P}$ denotes the set that two filter elements belong to the same pooling kernel,  $i_\eta$ and $j_\eta$ denote the indexes of elements in weight vector $w_d$. In Eq.~\ref{eq:optimization_primal},
${p_d}\in \mathbb{R}^N$ is a binary mask which crops the filter weights corresponding to the target region. By introducing ${p_d}$, we make sure that the filter only has the response for the target region of each circularly constructed sample~\cite{galoogahi2017learning}.
The vector $g_d\in \mathbb{R}^N$ is a regularization weight that encourages the filter to learn more weights in the central part of the target object.
The idea to introduce ${p_d}$ and $g_d$ has been previously proposed in~\cite{danelljan2015learning,galoogahi2017learning}, while our tracker is the first attempt to integrate them.
In the equality constraints, we consider the relationships between two arbitrary weight elements in a pooling kernel, thus $K = \frac{{{e}!}}{{({e} - 2)!2!}} (\left\lfloor {(L-e)/e} \right\rfloor  + 1) $ for each channel $d$, where $L$ is the number of nonzero values in $p_d$.
Note that the constraints are only performed in the filter coefficients corresponding to the target region of each sample, and the computed $K$ is based on the one-dimensional case.

According to the Parseval's formula, the optimization in Eq.~\ref{eq:optimization_primal} can be 
equivalently written as:
\begin{equation}
\begin{array}{l}
E(\hat w) =  \frac{1}{2}\left\| {\hat y - \sum\limits_d^D {{\hat P_d}{{\hat w}_d} \odot \hat x{}_d} } \right\|_2^2 +  \frac{\lambda}{2}\sum\limits_{d = 1}^D\left\| {{\hat G_d} {w_d}} \right\|_2^2 \\
{\rm s.t.}{\kern 1pt} {\kern 1pt} {\kern 1pt} V_d^1\mathcal{F}_d^{-1}{{\hat w}_d} = V_d^2\mathcal{F}_d^{-1}{{\hat w}_d}
\end{array},
\label{equ:Parseval}
\end{equation}
where $\mathcal{F}_d$ denotes the Fourier transform matrix, and $\mathcal{F}_d^{-1}$ denotes the inverse transform matrix.
The vectors $\hat p_d\in \mathbb{C}^{N\times 1}$, $\hat y\in \mathbb{C}^{N\times 1}$, $\hat x_d\in \mathbb{C}^{N\times 1}$ and $\hat w_d\in \mathbb{C}^{N\times 1}$ denote the Fourier coefficients of the corresponding signal vectors $y$, $x_d$, $p_d$ and $w_d$. 
Matrices $\hat P_d$ and $\hat G_d$ are the Toeplitz matrices, whose $(i,j)$-th elements are
$\hat p_d( (N + i-j)\%N + 1 )$ and $\hat g_d( (N+ i-j)\%N + 1)$, where $\%$ denotes the modulo operation. They are constructed based on the convolution theorem to ensure that $\hat P_d \hat w_d = \hat p_d * \hat w_d $, $\hat G_d w_d =\hat g_d* \hat w_d$.
Since the discrete Fourier coefficients of a real-valued signal are Hermitian symmetric, \ie, $\hat p_d( (N + i-j)\%N + 1 ) = {\hat p}_d( (N + j-i)\%N + 1 )^*$ in our case,  we can easily conclude that $\hat P_d = \hat P_d^H$ and $\hat G_d = \hat G_d^H$, where $H$ denotes the  conjugate-transpose of a complex matrix.
In the constraint term, $V_d^1\in \mathbb{R}^{K\times N}$ and $V_d^2\in \mathbb{R}^{K\times N}$ are index matrices with either $1$ or $0$ as the entries, $V_d^1\mathcal{F}_d^{-1}{{\hat w}_d} = [w_d(i_1),...,w_d(i_K)]^\top$ and  $V_d^2\mathcal{F}_d^{-1}{{\hat w}_d} = [w_d(j_1),...,w_d(j_K)]^\top$.

Eq.~\ref{equ:Parseval} can be rewritten in a compact formula as:
\begin{equation}
\begin{array}{l}
E(\hat w) =  \frac{1}{2}\left\| {\hat y -\sum\limits_{d = 1}^D {{\hat E}_d}{{\hat w}_d}} \right\|_2^2 +  \frac{\lambda}{2}\sum\limits_{d = 1}^D\left\| {{{\hat G}_d}{{\hat w}_d}} \right\|_2^2\\
{\rm s.t.}{\kern 10pt} {V_d}{\mathcal{F}_d^{ - 1}}{{\hat w}_d} = {\bf{0}}
\end{array},
\label{equ:simple_form}
\end{equation}
where $\hat E_d=\hat X_d\hat P_d$, $\hat X_d = {\rm diag}(\hat x_d(1),...,\hat x_d(N) )$ is a diagonal matrix, ${V_d} = V_d^1 - V_d^2$.

\begin{figure}
  \centering
  \includegraphics[width=1\linewidth]{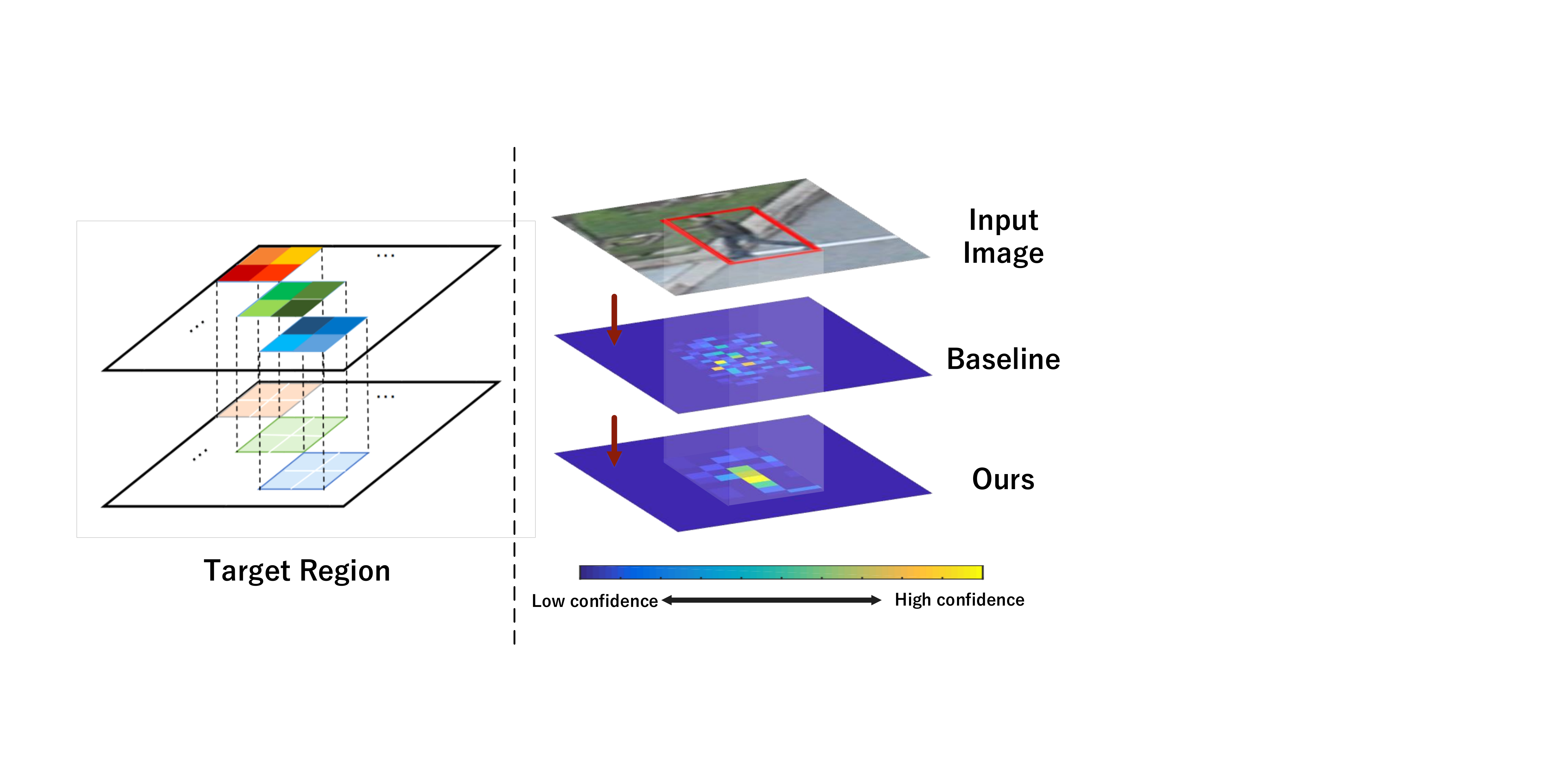}\\
  \PutCap
  \caption{Comparison between filter weights of the baseline method (\ie, the correlation filter algorithm without ROI-based pooling) and the proposed method.
  (a) A toy model showing that our learned filter elements are identical in each pooling kernel.
  (b) Visualizations of the filter weights learned by the baseline and our method. Our algorithm learns more compact filter weights than the baseline method, and thus can better address the over-fitting problem.
  }
  \label{fig:demo_pooling}
\end{figure}

%--------------------------------------------------------
\subsection{Model Learning}
Since Eq.~\ref{equ:simple_form} is a quadratic programming problem with linear constraints, we use the Augmented Lagrangian
Method for efficient model learning. The Lagrangian function corresponding to Eq.~\ref{equ:simple_form} is defined as:
\begin{equation}
\begin{array}{l}
\mathcal{L}(\hat w,\xi ) = \frac{1}{2}\left\| {\hat y - \sum\limits_{d = 1}^D {{{\hat E}_d}{{\hat w}_d}} } \right\|_2^2 + \frac{\lambda}{2} \sum\limits_{d = 1}^D {\left\| {{{\hat G}_d}}{\hat w_d} \right\|_2^2} \\
 + \sum\limits_{d = 1}^D {\xi _d^\top {V_d}{\mathcal{F}_d^{ - 1}}{{\hat w}_d}}  +  \frac{1}{2}\sum\limits_{d = 1}^D {{\gamma _d}\left\| {{ V_d}{\mathcal{F}_d^{ - 1}}{{\hat w}_d}} \right\|_2^2},
\end{array}
\label{equ:Lagrangian_gammaltiplier}
\end{equation}
where $\xi_d\in \mathbb{R}^K$ denotes the Lagrangian multipliers for the $d$-th channel, $\gamma_d$ is the penalty parameter, $\xi = [\xi_1^\top, ..., \xi_D^\top]^\top$. The ADMM method is used to alternately optimize $\hat w$ and $\xi$.
Though the optimization objective function is non-convex, it becomes a convex function when either $\hat w$ or $\xi$ is fixed.

When $\xi$ is fixed, $\hat w$ can be computed via the conjugate gradient descent method~\cite{bunse1999conjugate}. We compute the gradient of the objective function with respects to $\hat w_d$ in Eq.~\ref{equ:Lagrangian_gammaltiplier} and obtain a number of linear equations by setting the gradient to be a zero vector: 
\begin{equation}
(\hat A + \mathcal{F}{{\overline V }^\top} {\overline V}{\mathcal{F}^{ - 1}} + \lambda {{\hat G}^H}\hat G)\hat w = {{\hat E}^H}y - \mathcal{F}{V^\top}\xi ,
\label{equ:norm_equations}
\end{equation}
where $\mathcal{F}\in\mathbb{C}^{DN\times DN}$, $\hat G \in \mathbb{C}^{DN\times DN}$, $V\in \mathbb{R}^{DK\times DN}$ and $\overline V\in \mathbb{R}^{DK\times DN}$ are block diagonal matrices with the $d$-th matrix block set as $\mathcal{F}_d$, $\hat G_d$, $V_d$ and $\sqrt{\gamma_d}V_d$, $E = [{E_1},{E_2},...,{E_D}]$,
$\hat A = {E^H}E$.
In the conjugate gradient method,
the computation load lies in the three terms $\hat A \hat u$, $\mathcal{F}{{\overline V}^\top}{\overline V}{\mathcal{F}^{ - 1}}\hat u$ and $\lambda {{\hat G}^H}\hat G \hat u$ given the search direction $\hat u = [u_1^\top,...,u_D^\top]^\top$. In the following, we present more details on how we compute these three terms efficiently.
Each of the three terms can be regarded as a vector constructed with $D$ sub-vectors. The $d$-th sub-vector of $\hat A \hat u$ is computed as ${\hat P_d^HX_d^H\sum\limits_{j = 1}^D {{{\hat X}_j}({{\hat P}_j}{{\hat u}_j}}) }$ wherein $P_d^H=P_d$ as described above.
Since the Fourier coefficients of $p_d$ (a vector with binary values) are densely distributed, it is time consuming to directly compute $\hat P_d \hat v$ given an arbitrary complex vector $\hat v$. In this work, the convolution theorem is used to efficiently compute $\hat P_d \hat v$. The $d$-th sub-vector of the second term is ${\mathcal{F}_d}{\overline V_d}^\top{\overline V_d}{u_d} = \gamma_d {\mathcal{F}_d}{V_d}^\top{V_d}{u_d} $. As the matrices $V_d$ and $V_d^\top$ only consists of $1$ and $-1$, thus the computation of $V_d^\top{V_d}u_d$ can be efficiently conducted via table lookups. The third term corresponds to the convolution operation, whose convolution kernel is usually smaller than 5, thus it can also be efficiently computed. 

When $\hat w$ is computed, $\xi_d$ can be updated via:
\begin{equation}
    \xi _d^{i + 1} = \xi _d^i + {\gamma _d}{V_d}\mathcal{F}_d^{ - 1}{{\hat w}_d},
\end{equation}
where we use $\xi _d^{i}$ to denote the value of $\xi _d$ in the $i$-th iteration. According to~\cite{boyd2011distributed}, the value of $\gamma_d$ can be updated as:
\begin{equation}
    \gamma_d^{i+1} ={\rm min}(\gamma_{\rm max}, \alpha \gamma_d^i),
\end{equation}
again we use $i$ to denote the iteration index.

\subsection{Model Update}
To learn more robust filter weights, we update the proposed RPCF tracker based on several training samples ($T$ samples in total)  like~\cite{danelljan2016beyond,danelljan2017eco}. We extend the notations $\hat A$ and $\hat E$ in Eq.~\ref{equ:norm_equations} with superscript $t$, and reformulate Eq.~\ref{equ:norm_equations} as follows:
\begin{equation}
(\sum\limits_{t=1}^T\mu_t\hat A^t +  \mathcal{F}{{V}^\top}V{\mathcal{F}^{ - 1}} + \lambda {{\hat G}^H}\hat G)\hat w =b,
\end{equation}
where $b=\sum\limits_{t=1}^T \mu_t{(\hat E^t)}^Hy - \mathcal{F}{V^\top}\xi $, and 
$\mu_t$ denotes the importance weight for each training sample $t$. 
Most previous correlation filter trackers update the model iteratively via a weighted combination of the filter weights in various frames. 
Different from them, we exploit the sparse update mechanism, and update the model every $N_t$ frames~\cite{danelljan2017eco}. In each updating frame, the conjugate gradient descent method is used, and the search direction of the previous update process is input as a warm start. Our training samples are generated following~\cite{danelljan2017eco}, and the weight (\ie, learning rate) for the newly added sample is set as $\omega $, while the weights of previous samples are decayed by multiplying $1-\omega$. In Figure~\ref{fig:demo_pooling}, we visualize the learned filter weights of different trackers with and without ROI-based pooling, our tracker can learn more compact filter weights and focus on the reliable regions of the target object.
%-------------------------------------------------------------------------

\subsection{Target Localization}
In the target localization process, we first crop the candidate samples with different scales, \ie, $x_d^s, s\in\{1,...,S\}$. Then, we compute 
the response $\hat r^s$ for the feature in each scale in the Fourier domain:
\begin{equation}
    {\hat r^s} = \sum\limits_{d = 1}^D {\hat x_d^s} {\hat w_d}.
\end{equation}

The computed responses are then interpolated with trigonometric polynomial following~\cite{danelljan2015learning} to achieve the sub-pixel target localization.

%-------------------------------------------------------------------------
\section{Experiments}
\label{sec:experiments}
In this section, we evaluate the proposed RPCF tracker on the OTB-2013~\cite{wu2013online}, OTB-2015~\cite{wu2015object} and VOT2017~\cite{VOT2017} datasets. We first evaluate the effectiveness of the method, and then further compare our tracker with the recent state-of-the-art.

\subsection{Experimental Setups}
\noindent\textbf{Implementation Details.} The proposed RPCF method is mainly implemented in MATLAB on a PC with an i7-4790K CPU and a Geforce 1080 GPU.
Similar to the ECO method~\cite{danelljan2017eco}, we use a combination of CNN features from two convolution layers, HOG and color names for target representation. For efficiency, the PCA method is used to compress the features. We set the learning rate $\omega $, the maximum number of training samples $T$, $\gamma_{\rm max}$ and $\alpha$ as 0.02, 50,  1000 and 10 respectively, and we update the model in every $N_t$ frame.
As to $\gamma_d$, we set a relative small value $\gamma_1$ (\eg,~$0.1$) for the high-level feature (\ie, the second convolution layer), and a larger value $\gamma_2 = 3\gamma_1$ for the other feature channels. The kernel size $e$ is set as $2$ in the implementation. We use the conjugate gradient descent for model initialization and update, 200 iterations are used in the first frame, and the following update frame uses 6 iterations. 
Our tracker runs at about 5fps without optimization.

\noindent\textbf{Evaluation Metric.}
We follow the one-pass evaluation (OPE) rule on the OTB-2013 and OTB-2015 datasets, and report the precision plots as well as the success plots for the performance measure. The success plots demonstrate the overlaps between tracked bounding boxes and ground truth with varying thresholds, while the precision plots measure the accuracy of the estimated target center positions. In the precision plots, we exploit the distance precision (DP) rate at 20 pixels for the performance report, while we exploit the area-under-curve (AUC) score for performance report in success plots.
On the VOT-2017 dataset, we evaluate our tracker in terms of the Expected Average Overlap (EAO), accuracy raw value (A) and robustness raw value (R) measure the overlap, accuracy and robustness respectively.
% The EAO value measures the reset-based overlap of a tracker. The A and R measure the performance of different trackers on the accuracy and robustness respectively.

\begin{figure}[h]
\centering
\begin{tabular}{c@{}c}
\includegraphics[scale=.28]{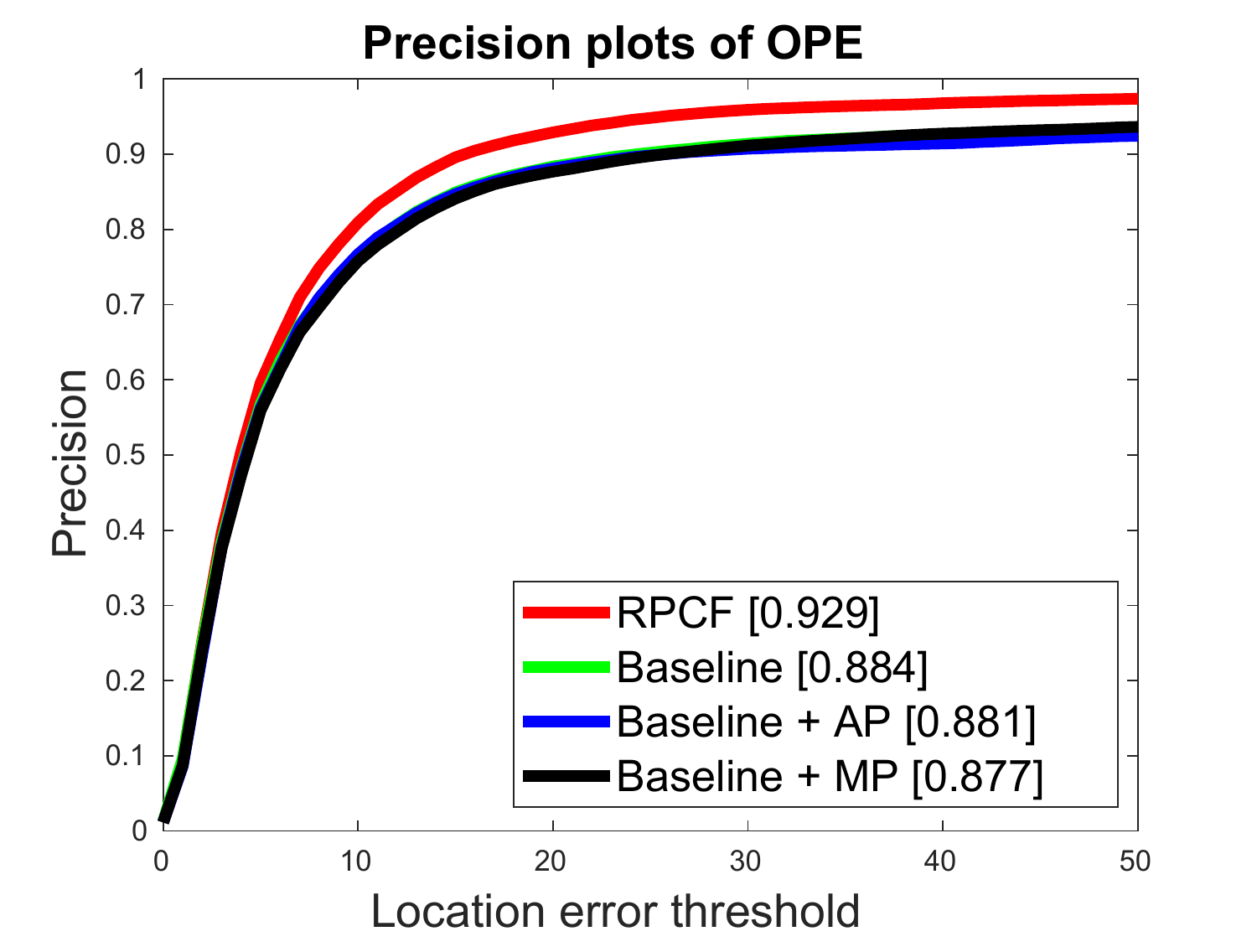}
\ & 
\includegraphics[scale=.28]{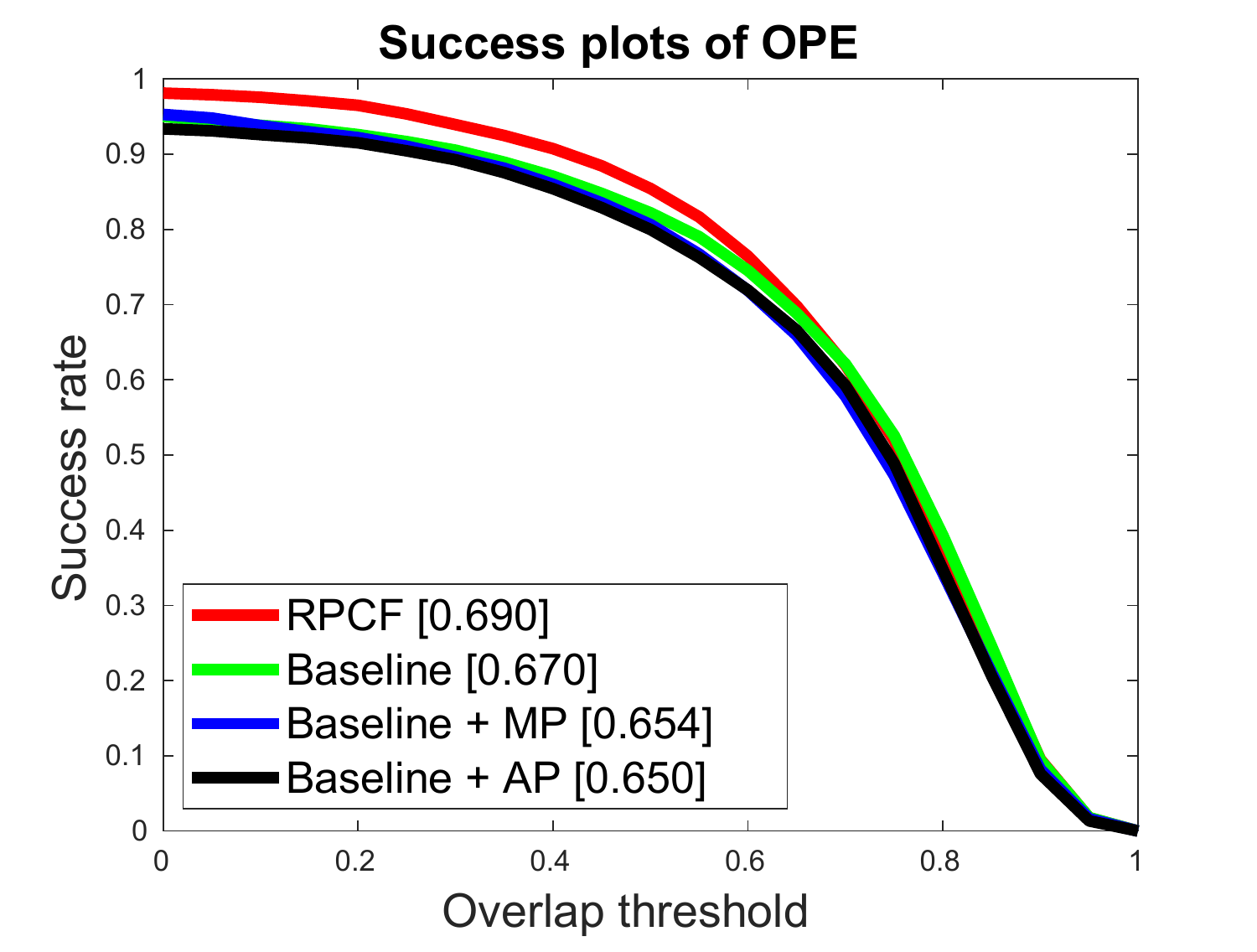}
\ \\
(a) & (b)
\end{tabular}

\caption{ Precision and success plots of 100 sequences on the OTB-2015 dataset. The distance precision rate at the threshold of 20 pixels and the AUC score for each tracker is presented in the legend.}
\label{fig:ablation}
\end{figure}

% \begin{figure}
%   \centering
%   % Requires \usepackage{graphicx}
%   \includegraphics[width=0.7\linewidth]{pic/ablation.pdf}\\
%   \caption{ Performance evaluation of ablation study with baseline and RPCF-NC. Success plot over all 100 sequences using one-pass evaluation on the OTB-2015 benchmark.
%   }
%   \label{fig:ablation}
% \end{figure}

\subsection{Ablation Study}
In this subsection, we conduct experiments to validate the contributions of the proposed RPCF method. We set the tracker that does not consider the pooling operation as the baseline method, and use Baseline to denote it. It essentially corresponds to Eq.~\ref{eq:optimization_primal} without equality constraints. To validate the superiority of our ROI-based pooling method over feature map based average-pooling and max-pooling, we also implement the trackers that directly performs average-pooling and max-pooling on the input feature map, which are named as Baseline+AP and Baseline+MP.

We first compare the Baseline method with Baseline+AP and Baseline+MP, which shows that the tracking performance decreases when feature map based pooling operations are performed.
Directly performing pooling operations on the input feature map will not only influence the extraction of the training samples but also lead to worse target localization accuracy. In addition, the over-fitting problem is not well addressed in such methods since the ratio between the numbers of model parameters and available training samples do not change compared with the Baseline method. We validate the effectiveness of the proposed method by comparing our RPCF tracker with the Baseline method. Our tracker improves the Baseline method by 4.4\% and 2.0\% in precision and success plots respectively. By exploiting the ROI-based pooling operations , our learned filter weights are insusceptible to the over-fitting problem and are more robust to deformations.

\begin{figure}[t]
\centering
\subfigure[]{
\includegraphics[scale=.27]{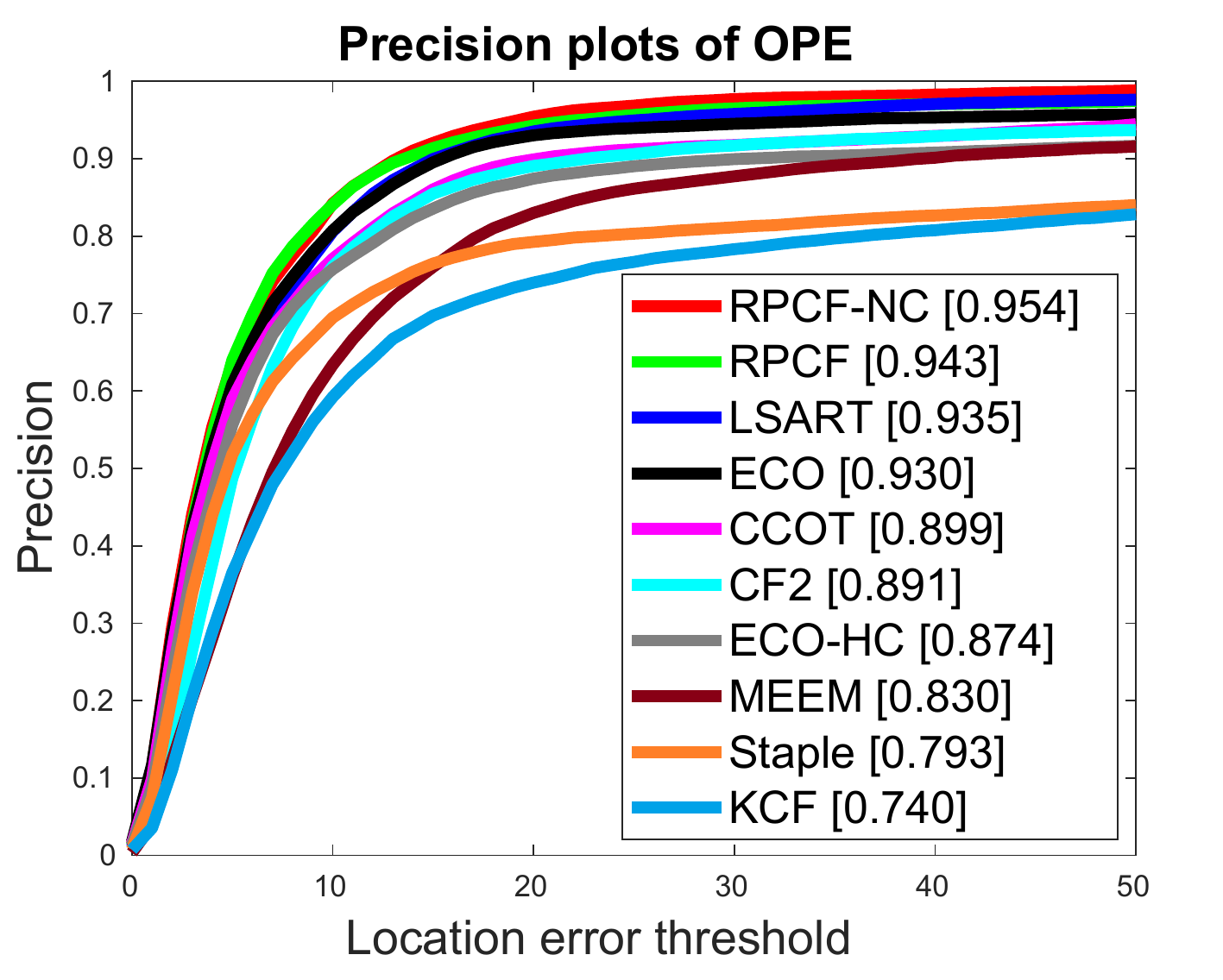}}
\subfigure[]{
\includegraphics[scale=.27]{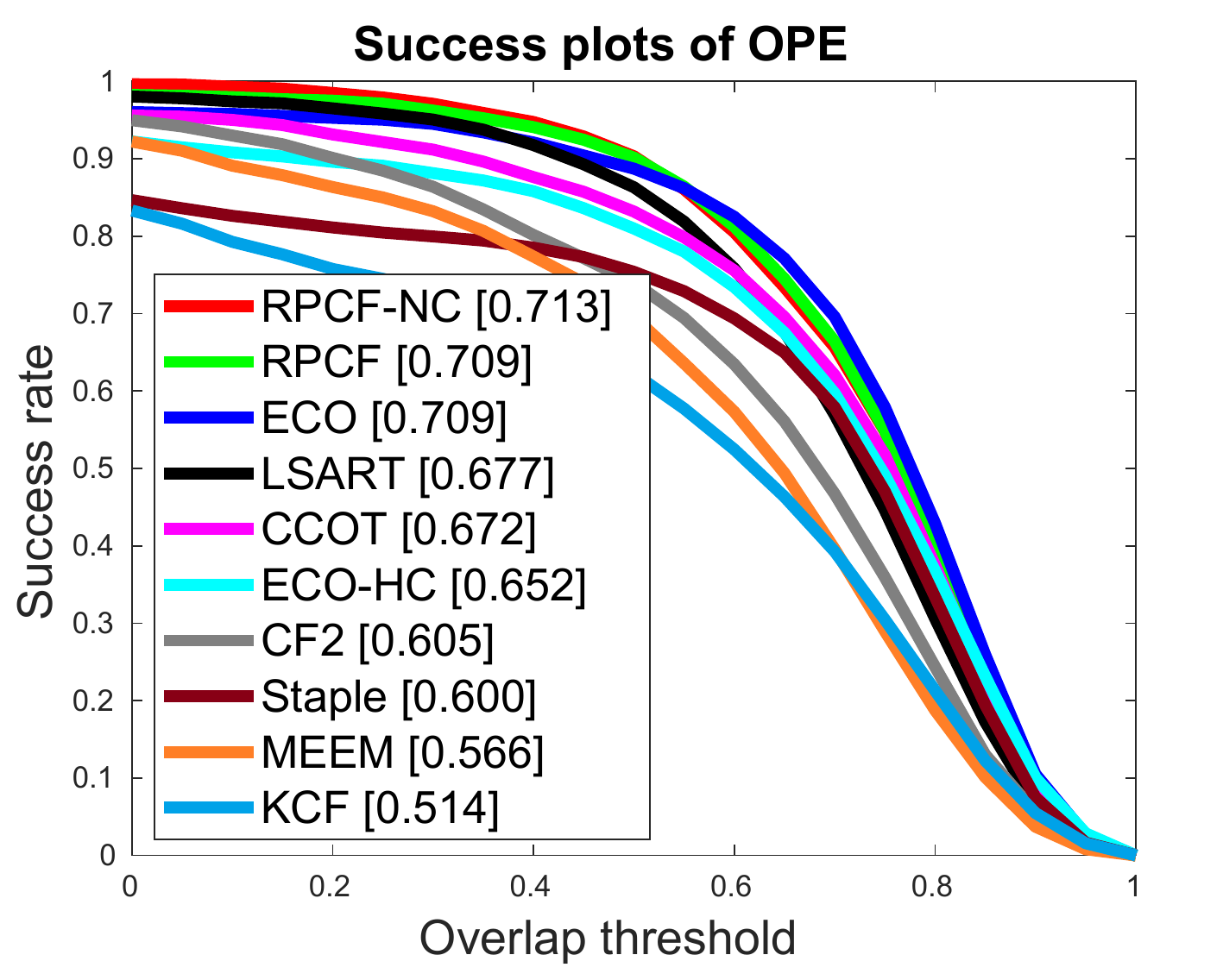}
}
\caption{ Precision and success plots of 50 sequences on the OTB-2013 dataset. The distance precision rate at the threshold of 20 pixels and the AUC score for each tracker is presented in the legend.}
\label{fig:otb-2013}
\end{figure}

\begin{figure}[t]
\centering
\subfigure[]{
\includegraphics[scale=.27]{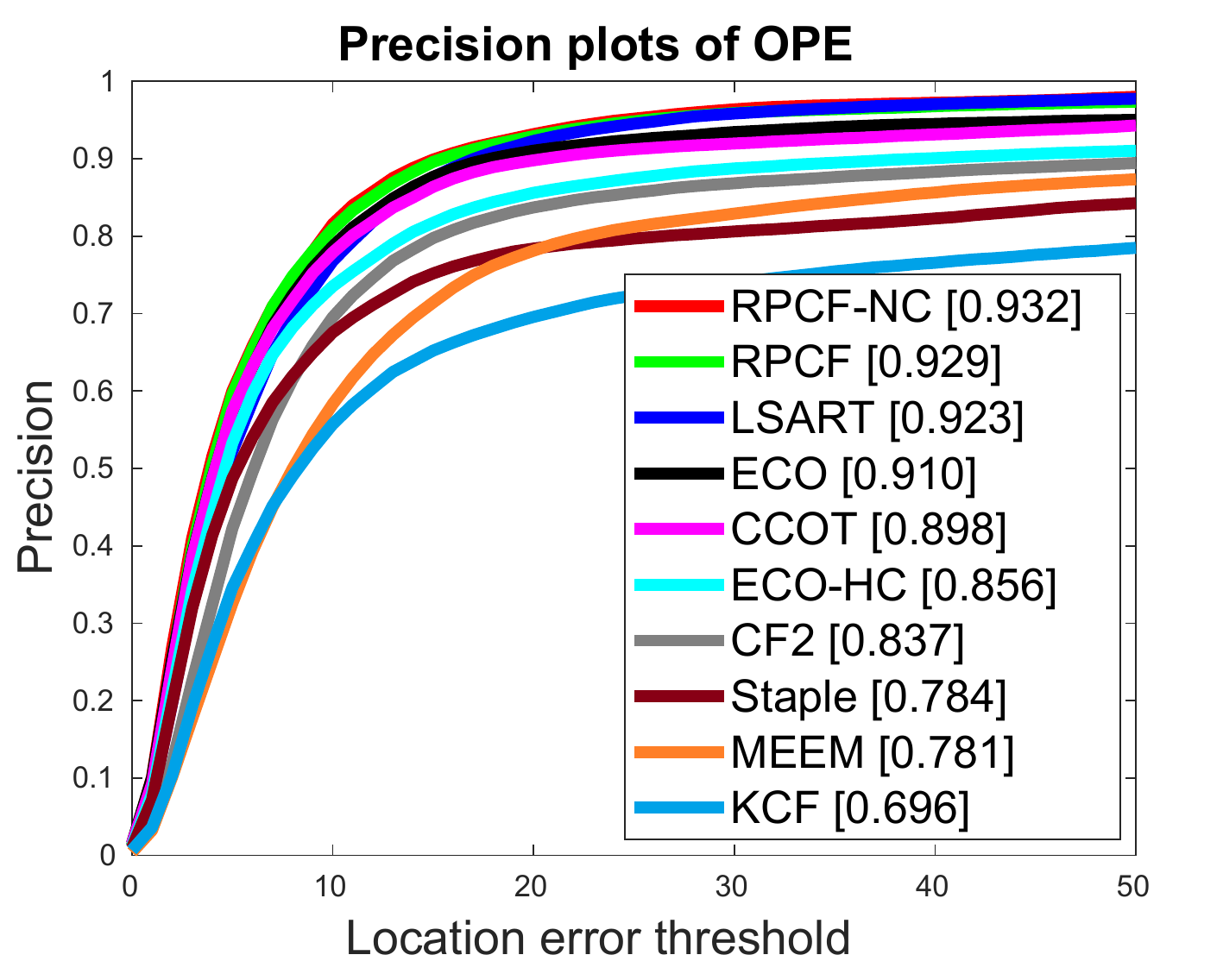}}
\subfigure[]{
\includegraphics[scale=.27]{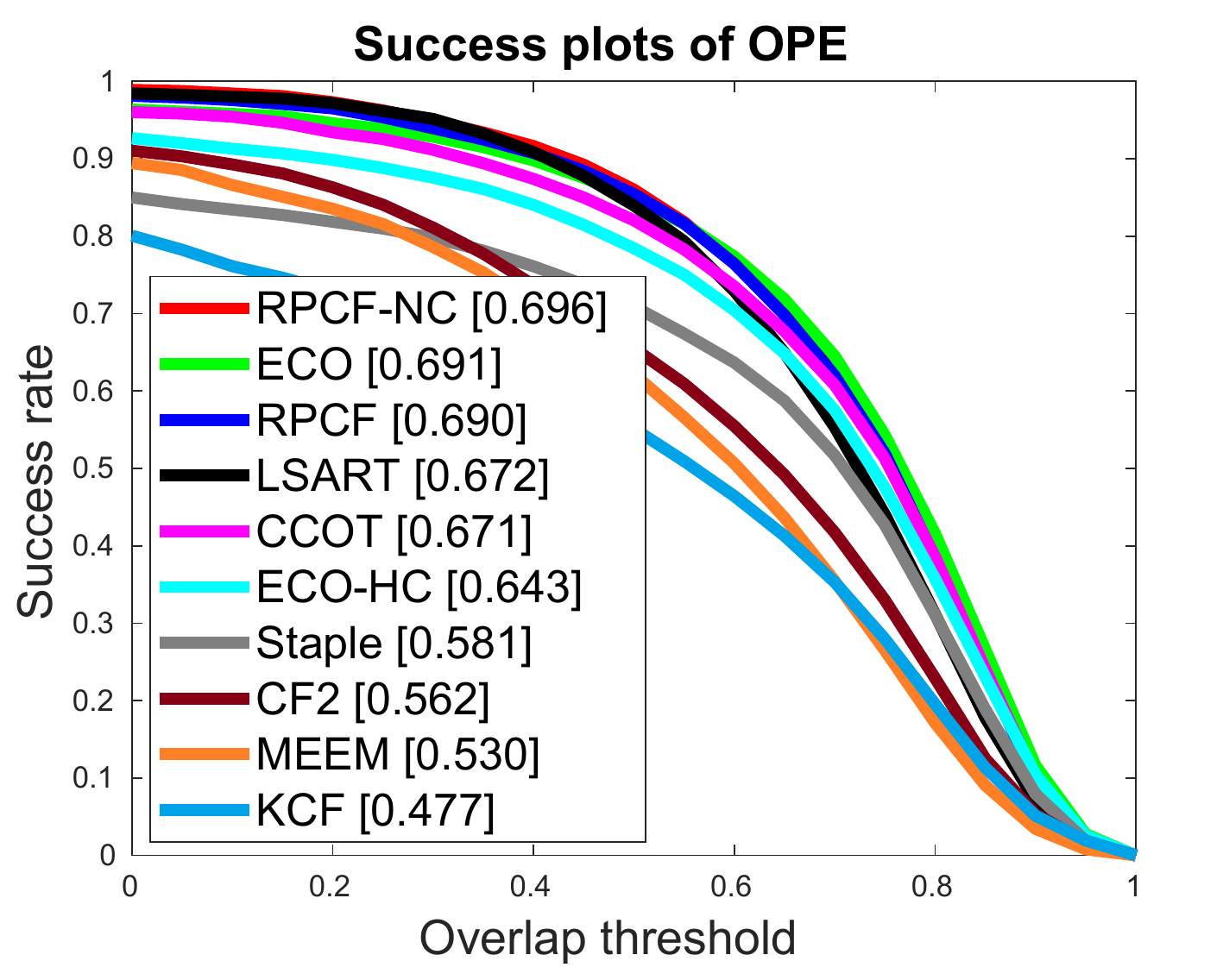}
}
\caption{ Precision and success plots of 100 sequences on the OTB-2015 dataset. The distance precision rate at the threshold of 20 pixels and the AUC score for each tracker is presented in the legend.}
\label{fig:otb-2015}
\end{figure}

\begin{figure}[h]
    \centering
    \includegraphics[width=0.9\linewidth, height = 47mm]{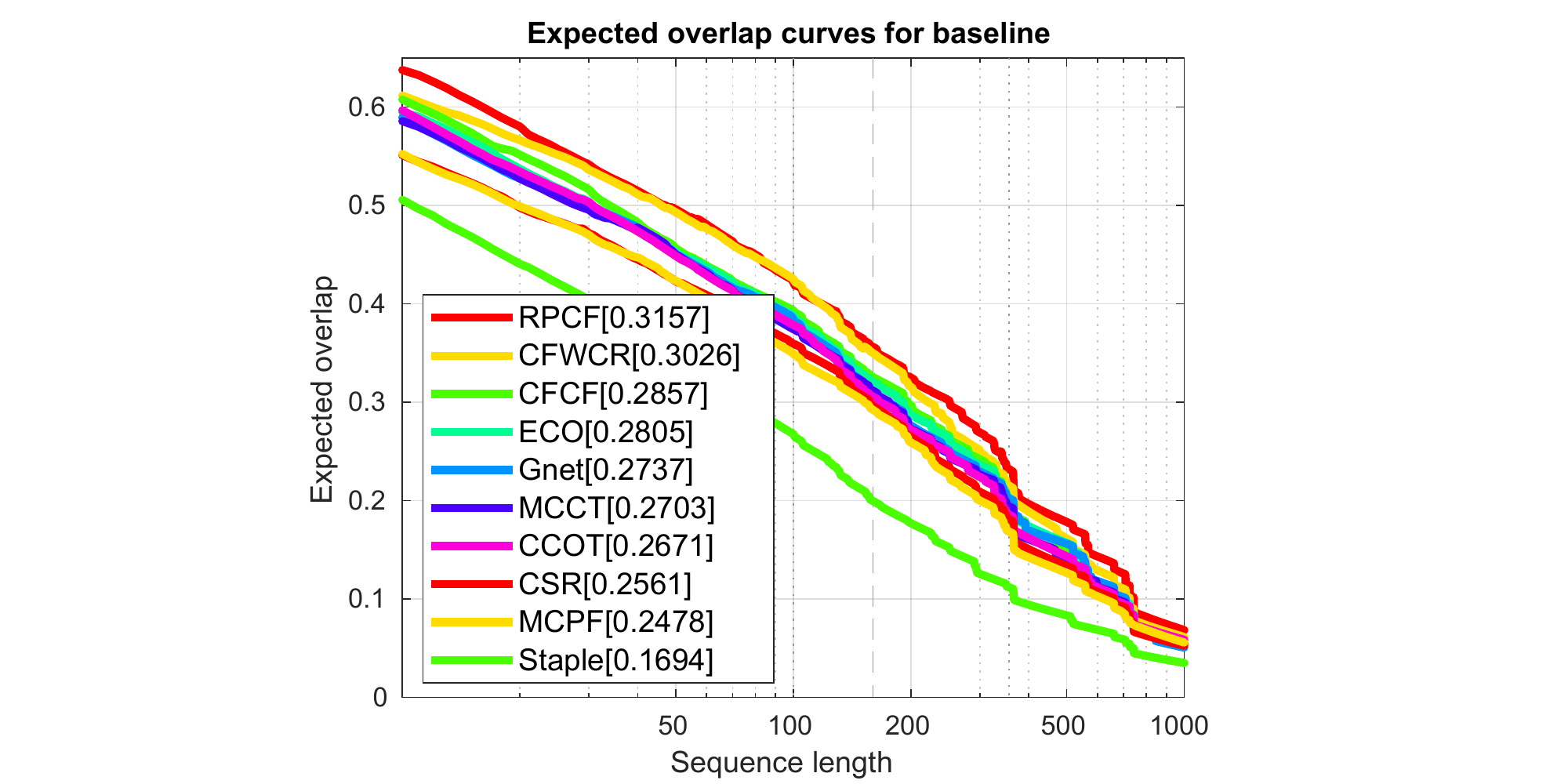}\\
    \caption{Expected Average Overlap (EAO) curve for 10 state-of-the-art
trackers on the VOT-2017 dataset.}
    \label{fig:vot-plot}
\end{figure}

\begin{figure*}[t!]
    \begin{center}
         \begin{tabular}{@{}c@{} @{}c@{} @{}c@{}  @{}c@{} }
     
          \includegraphics[scale=.27]{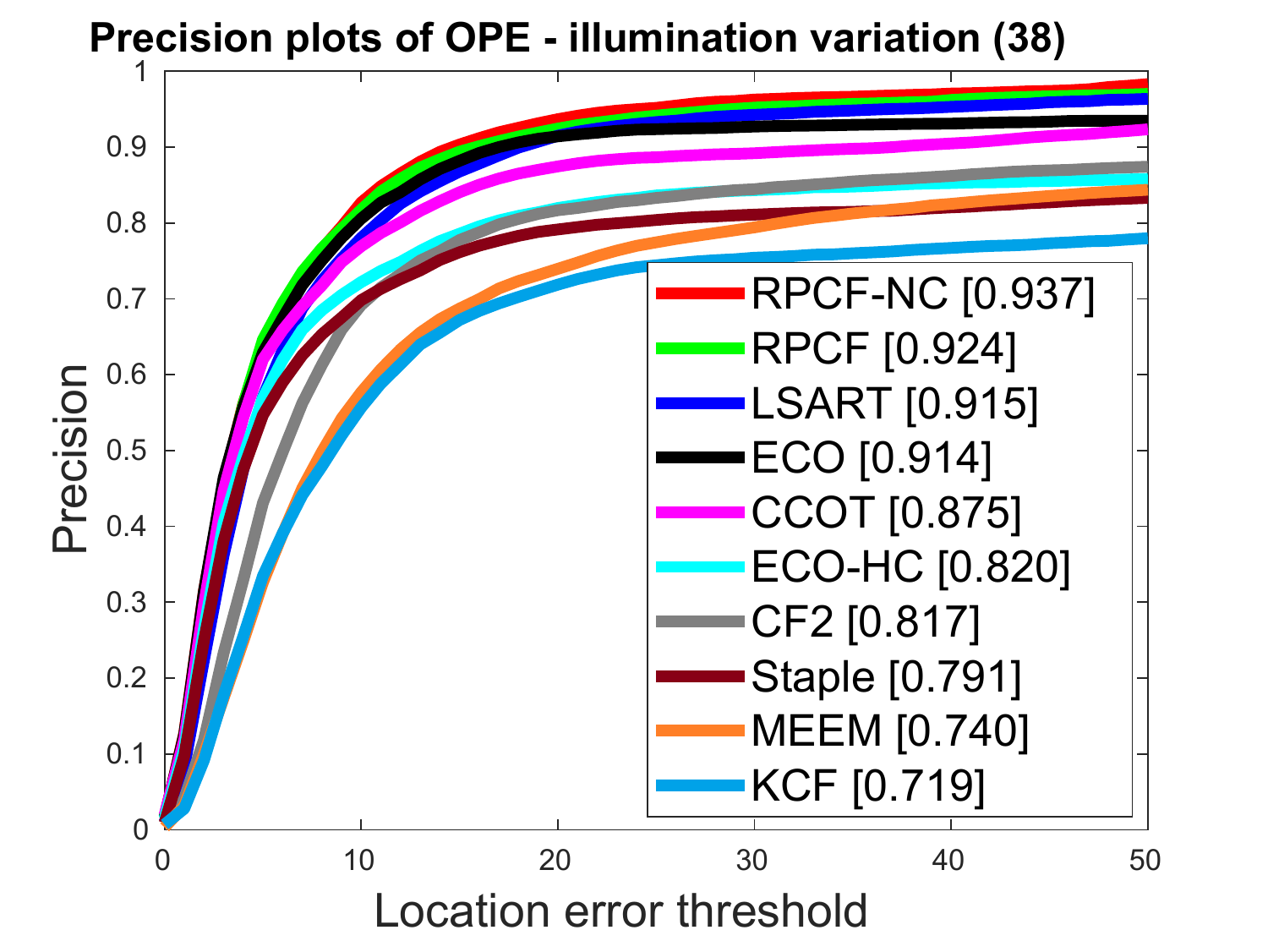}
&  \includegraphics[scale=.27]{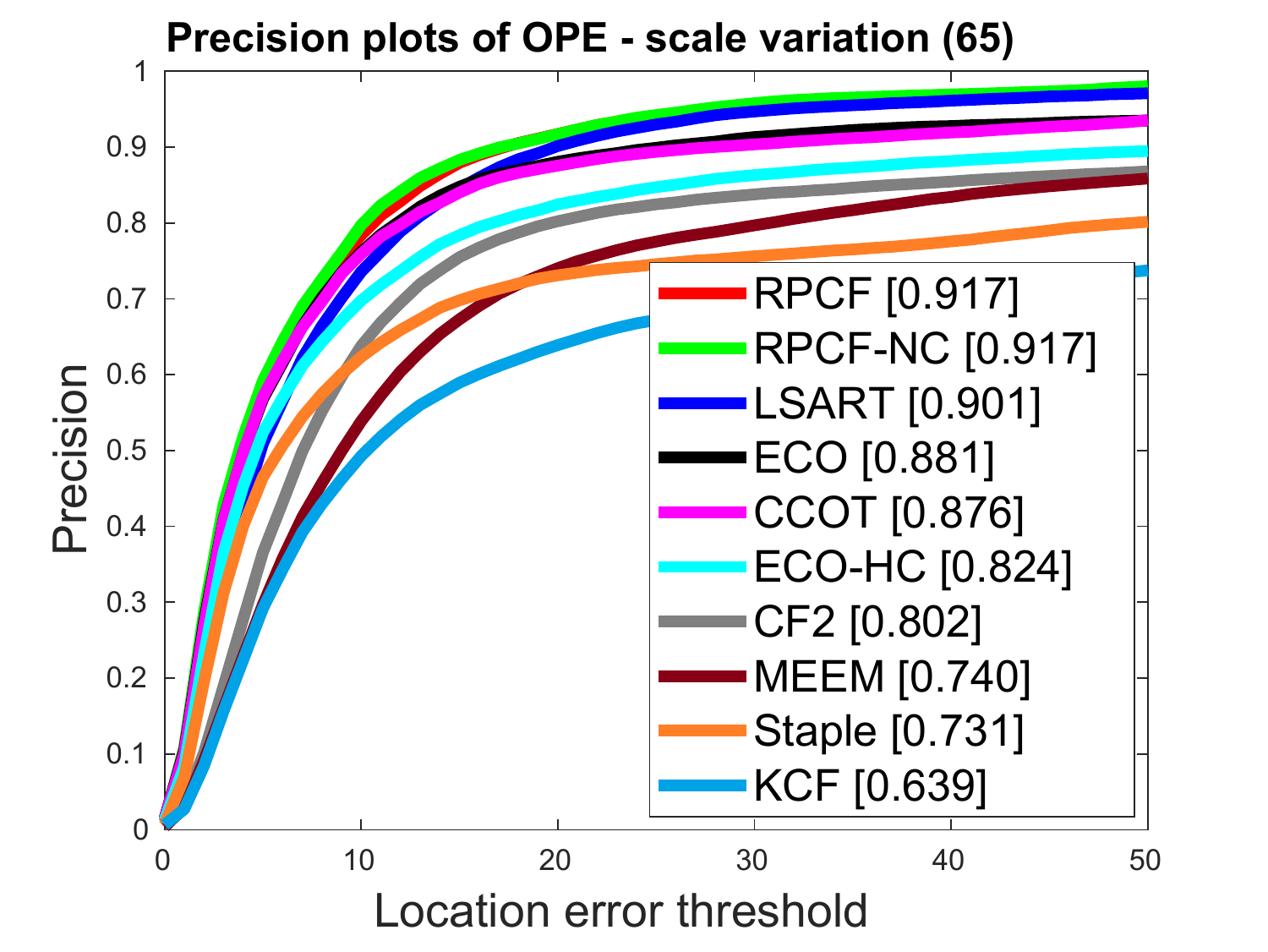}
&  \includegraphics[scale=.27]{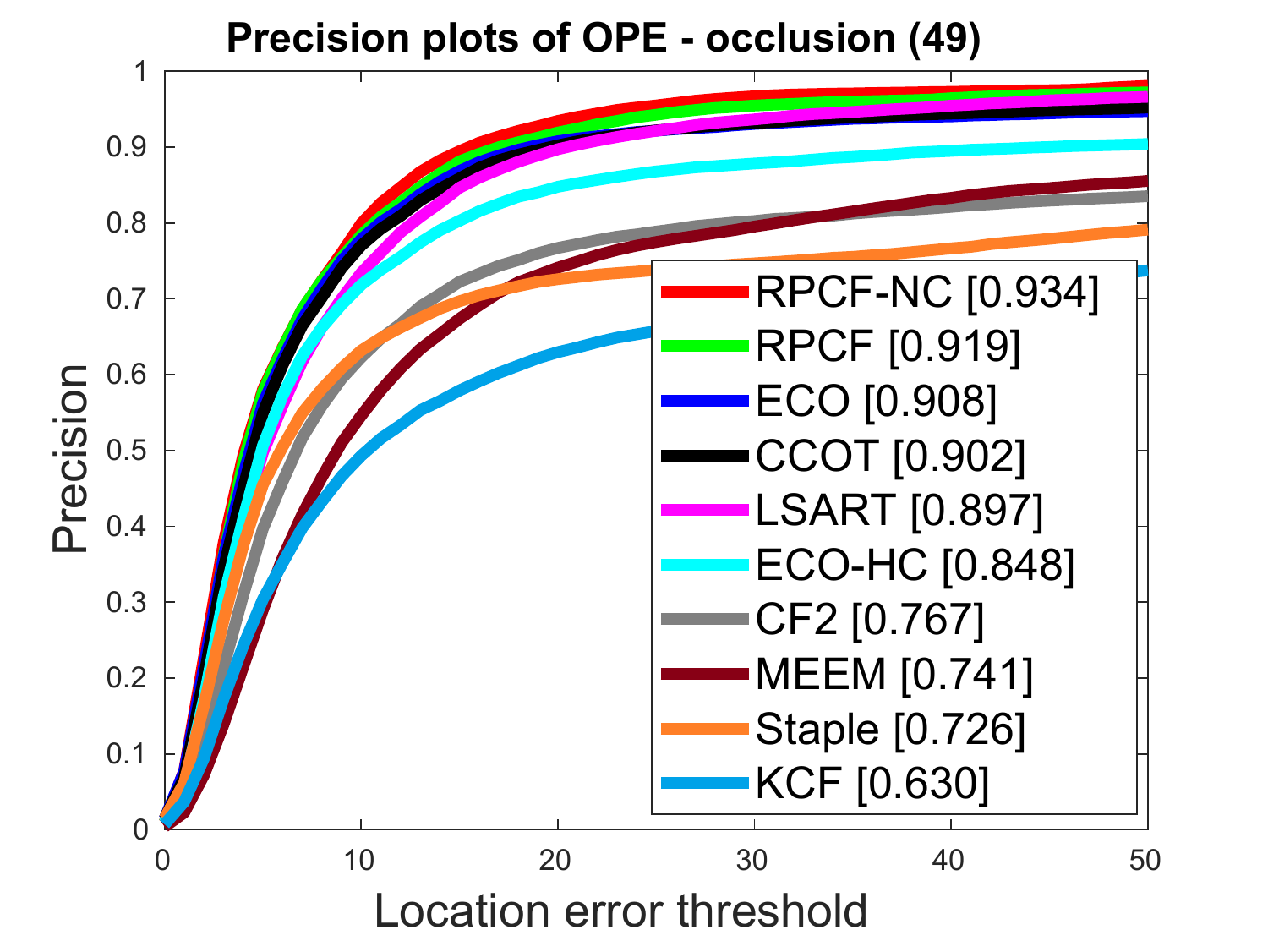}
& \includegraphics[scale=.27]{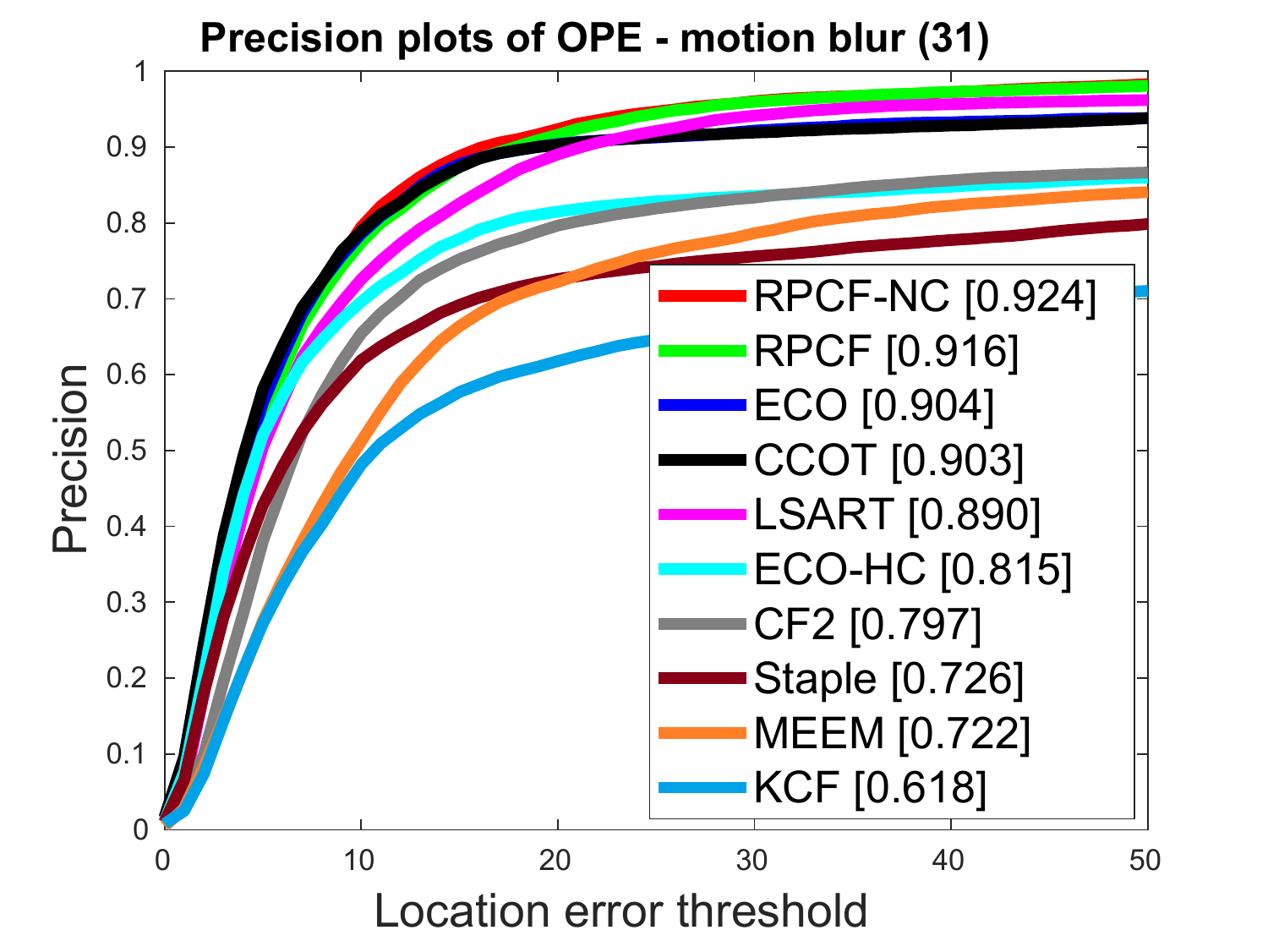}
  \\

    \includegraphics[scale=.27]{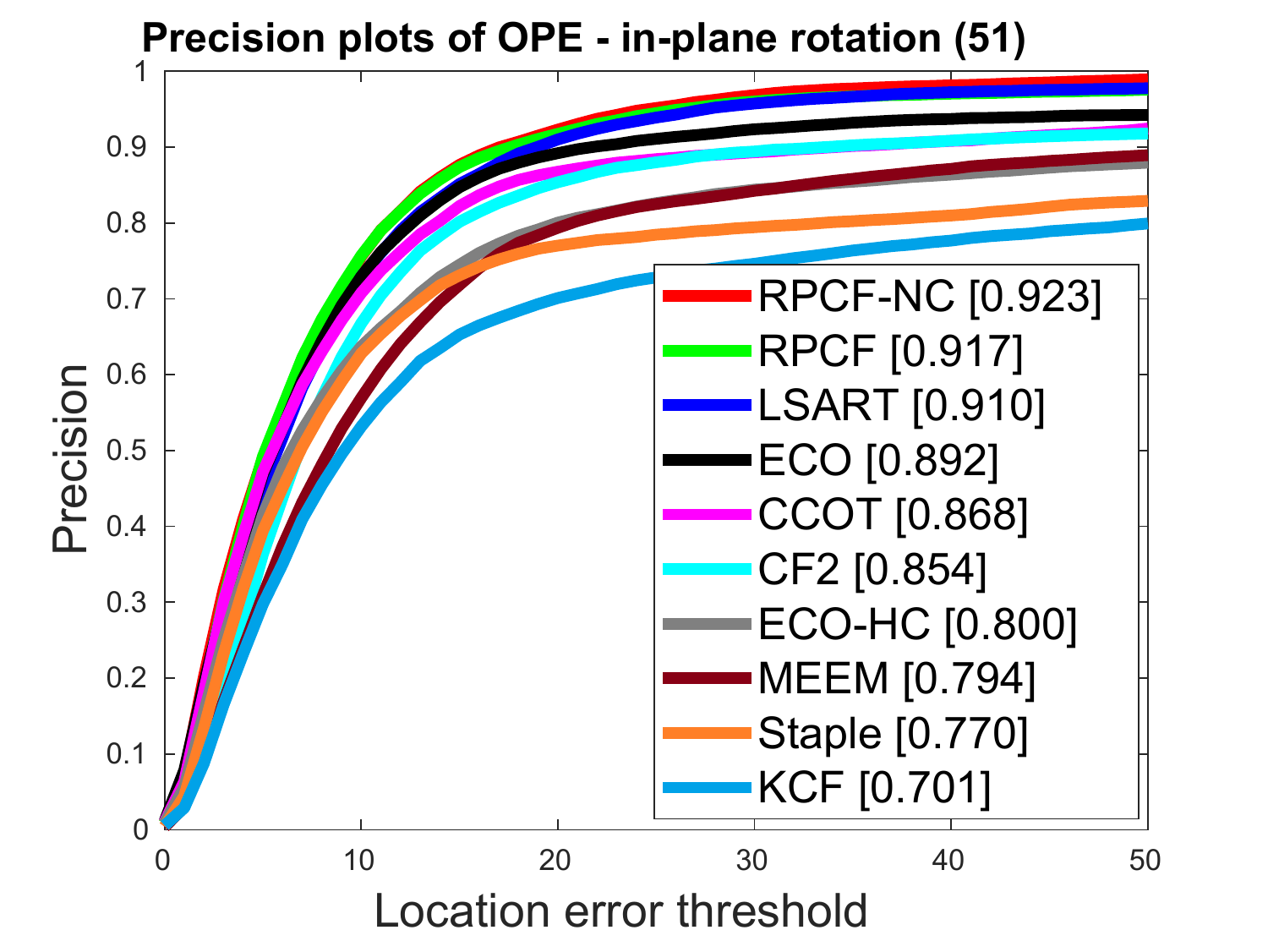}
&  \includegraphics[scale=.27]{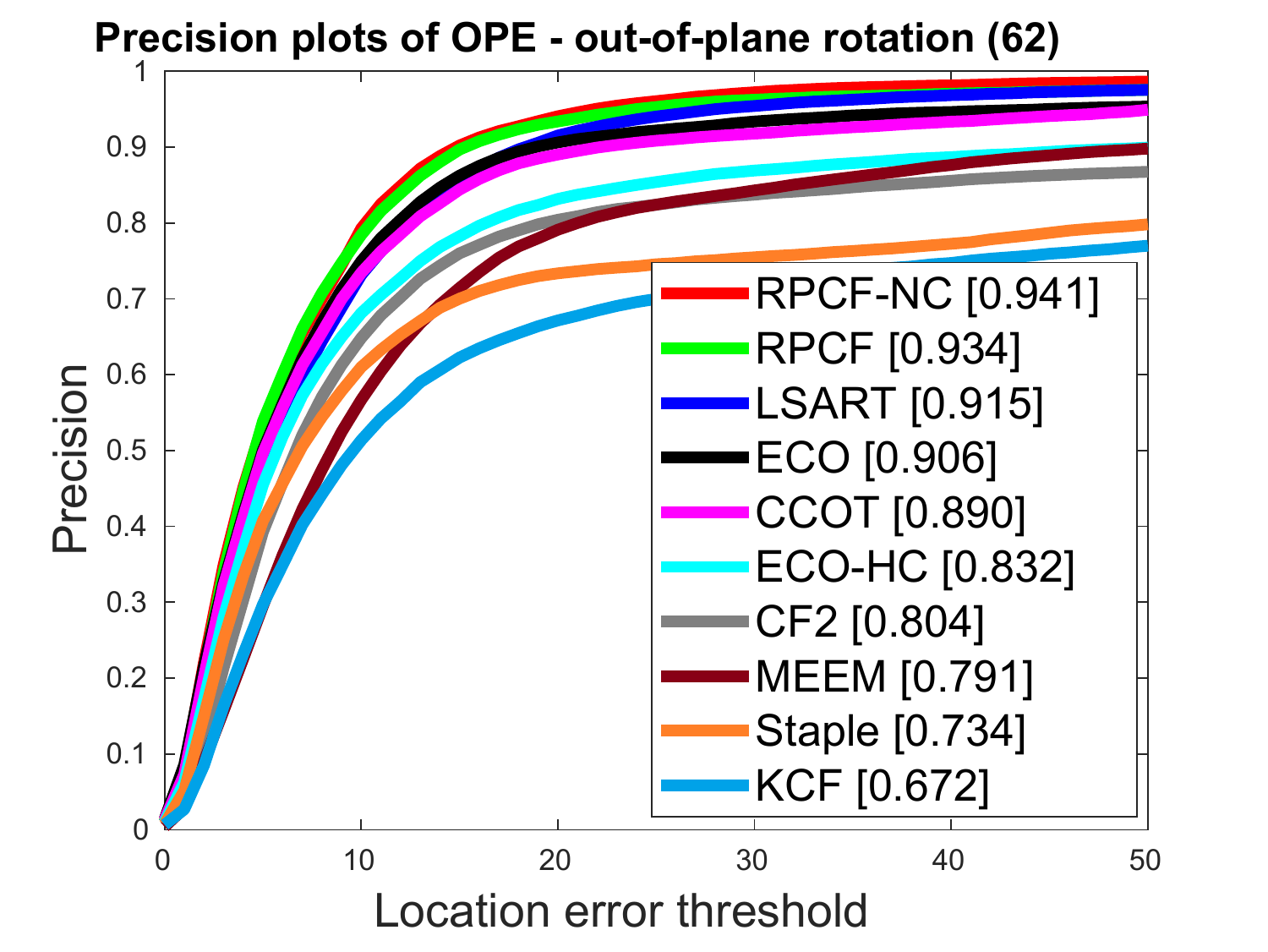}
&  \includegraphics[scale=.27]{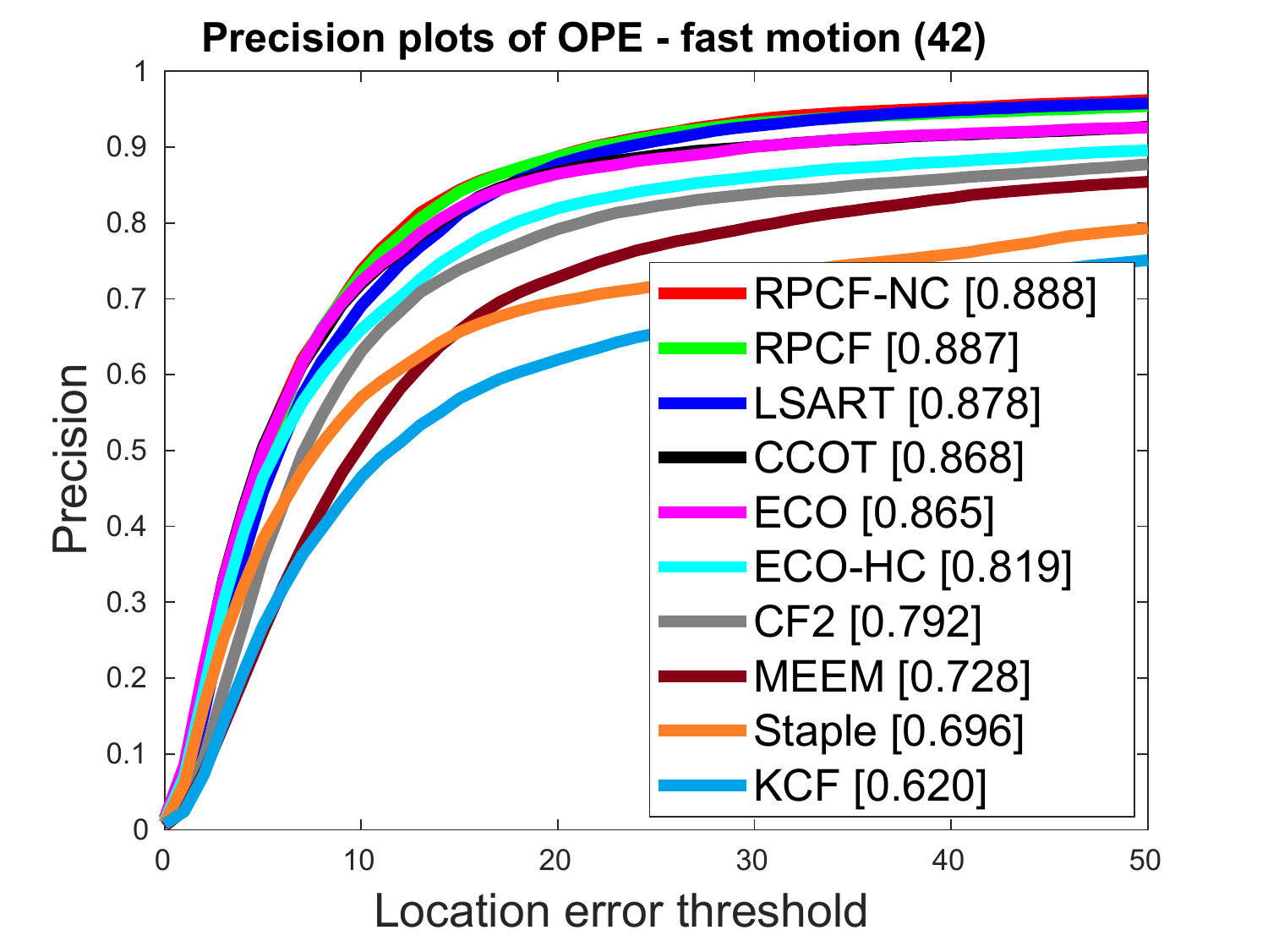} 
& \includegraphics[scale=.27]{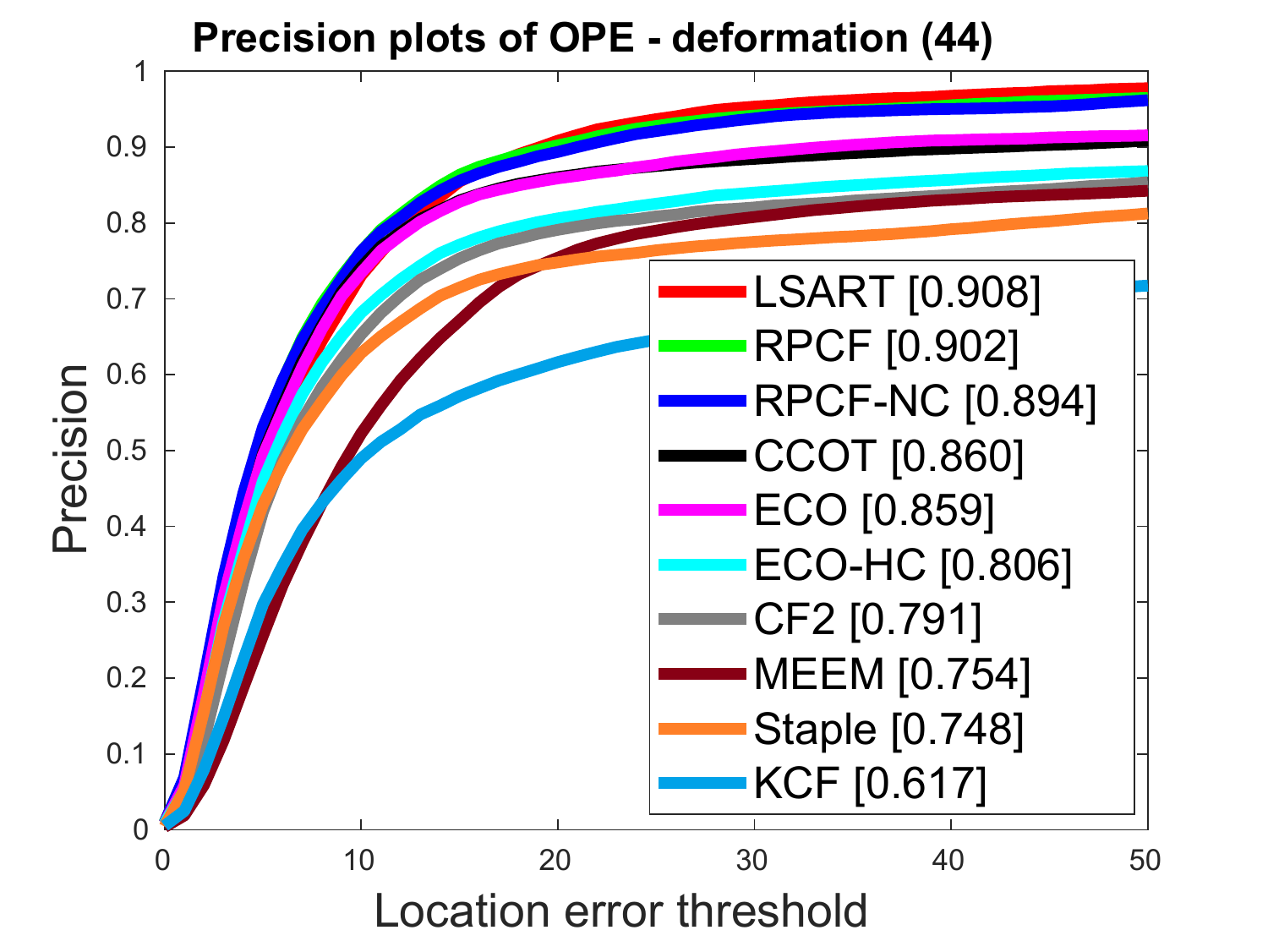}

             \end{tabular}
     \end{center}
            \caption{Precision plots of different algorithms on 8 attributes, which are respectively illumination variation, scale variation, occlusion, motion blur, in-plane rotation, out-of-plane rotation, fast motion and deformation.}
\label{Fig:Attr_OTB_100}
\end{figure*}

\begin{table*}[t!]
\centering
\caption{ Performance evaluation for 10 state-of-the-art algorithms on the VOT-2017 public dataset. The best three results are marked in red, blue and green fonts, respectively.}
\label{table:VOT-compare}
\begin{tabular}{lcccccccccc}
\hline
         & {\color{red}RPCF} & CFWCR  & CFCF & ECO & Gnet & MCCT & CCOT & CSR & MCPF & Staple \\ \hline
EAO  & {\color{red}0.316} & {\color{blue}0.303} & {\color{green}0.286} & 0.281 & 0.274 & 0.270 & 0.267 & 0.256 & 0.248 & 0.169 \\
A & 0.500 & 0.484 & 0.509 & 0.483 & 0.502 & {\color{blue}0.525} & 0.494 & 0.491 & {\color{green}0.510} & {\color{red}0.530}\\
R & {\color{red}0.234} & {\color{blue}0.267} & 0.281 & {\color{green}0.276} & {\color{green}0.276} & 0.323 & 0.318 & 0.356 & 0.427 & 0.688\\ \hline
% Ar & 2.62 &  2.80 &  2.42 & 2.50 & 2.43 & {\color{red}1.90} & 2.70 & 3.45 &  {\color{green}2.38} & {\color{blue}2.00}\\ 
% Rr  & {\color{red}2.48} & {\color{blue}2.57} & {\color{green}2.53} &  2.70  &  2.88  &  2.83  &  2.72  &  3.45  &   4.53  &   5.43 \\ \hline
\end{tabular}
\end{table*}

\subsection{State-of-the-art Comparisons}
\noindent\textbf{OTB-2013 Dataset.} The OTB-2013 dataset contains 50 videos annotated with 11 various attributes including illumination variation, scale variation, occlusion, deformation and so on. We evaluate our tracker on this dataset and compare it with 8 state-of-the-art methods that are respectively ECO~\cite{danelljan2017eco}, CCOT~\cite{danelljan2016beyond}, LSART~\cite{sun2018learning}, ECO-HC~\cite{danelljan2017eco}, CF2~\cite{ma2015hierarchical}, Staple~\cite{bertinetto2016staple}, MEEM~\cite{zhang2014meem} and KCF~\cite{henriques2015high}. 
We demonstrate the precision and success plots for different trackers in Figure~\ref{fig:otb-2013}.
Our RPCF method has a 94.3\% DP rate at the threshold of 20 pixels and a 70.9\%  AUC score.
Compared with other correlation filter based trackers, the proposed RPCF method has the best performance in terms of both precision and success plots. Our method improves the second best tracker ECO by 1.9\% in terms of DP rates, and has comparable performance according to the success plots. When the features are not compressed via PCA, the tracker (denoted as RPCF-NC) has a 95.4\% DP rate at the threshold of 20 pixels and a 71.3\% AUC score in success plots, and it runs at 2fps without optimization.

\textbf{OTB-2015 Dataset.}
The OTB-2015 dataset is an extension of the OTB-2013 dataset and contains 50 more video sequences. On this dataset, we also compare our tracker with the above mentioned 8 state-of-the-art trackers, and present the results in Fiugre~\ref{fig:otb-2015}(a)(b).
Our RPCF tracker has a 92.9\% DP rate and a 69.0\% AUC score. It improves the second best tracker ECO by 1.9\% in terms of the precision plots.
With the non-compressed features, our RPCF-NC tracker achieves the 93.2\% DP rate and 69.6\% AUC score, which again has the best performance among all the compared trackers.

The OTB-2015 dataset divides the image sequences into 11 attributes, each of which corresponds to a challenging factor.
We compare our RPCF tracker against other 8 state-of-the-art trackers and present the precision plots for different trackers in Figure~\ref{Fig:Attr_OTB_100}. As is illustrated in the figure, our RPCF tracker has good tracking performance in all the listed attributes.
Especially, the RPCF tracker improves the ECO method by 3.6\%, 2.5\%, 2.8\%,
2.2\% and 4.3\% in the attributes of scale variation, in-plane rotation, out-of-plane rotation, fast motion and deformation. The ROI pooled features become more consistent across different frames than the original ones, which contributes to robust target representation when the target appearance dramatically changes (see Figure~\ref{fig:pooling_robust} for example). In addition, by exploiting the ROI-based pooling operations, the model parameters are greatly compressed, which makes the proposed tracker insusceptible to the over-fitting problem. In Figure~\ref{Fig:Attr_OTB_100}, we also present the results of our RPCF-NC tracker for reference.

% \begin{tabular}{|c|c|c|c|c|}
% \hline
% \multirow{2}{*}{\textbf{ }} & \multicolumn{2}{ c }{\textbf{baseline}} & \multicolumn{2}{ c }{\textbf{Overall}} \\\hline
% \textbf{Accuracy} & \textbf{Robustness} & \textbf{Accuracy} & \textbf{Robustness} \\\hline
% \textbf{RPCF} & 2.62 & \first{2.48} & 2.62 & \first{2.48} \\\hline
% \textbf{CFWCR} & 2.80 & \third{2.57} & 2.80 & \third{2.57} \\\hline
% \textbf{CFCF} & 2.42 & \second{2.53} & 2.42 & \second{2.53} \\\hline
% \textbf{ECO} & 2.50 & 2.70 & 2.50 & 2.70 \\\hline
% \textbf{Gnet} & 2.43 & 2.88 & 2.43 & 2.88 \\\hline
% \textbf{MCCT} & \first{1.90} & 2.83 & \first{1.90} & 2.83 \\\hline
% \textbf{CCOT} & 2.70 & 2.72 & 2.70 & 2.72 \\\hline
% \textbf{CSR} & 3.45 & 3.45 & 3.45 & 3.45 \\\hline
% \textbf{MCPF} & \third{2.38} & 4.53 & \third{2.38} & 4.53 \\\hline
% \textbf{Staple} & \second{2.00} & 5.43 & \second{2.00} & 5.43 \\\hline
% \end{tabular}

\textbf{VOT-2017 Dataset.}
We test the proposed tracker on the VOT-2017 dataset for more thorough
performance evaluations.
The VOT-2017 dataset consists of 60 sequences with 5 challenging attributes, \ie, occlusion, illumination change, motion change, size change, camera motion. Different from the OTB-2013 and OTB-2015 datasets, it focuses on evaluating the short-term tracking performance and introduces a reset based experiment setting.
We compare our RPCF tracker with 9 state-of-the-art trackers including CFWCR~\cite{he2017correlation}, ECO~\cite{danelljan2017eco},
CCOT~\cite{danelljan2016beyond}, MCCT~\cite{wang2018multi}, CFCF~\cite{gundogdu2018good}, CSR~\cite{lukezic2017discriminative}, MCPF~\cite{zhang2017multi}, Gnet~\cite{VOT2017} and Staple~\cite{bertinetto2016staple}. 
The tracking performance of different trackers in terms of EAO, A and R are provided in Table~\ref{table:VOT-compare} and Figure~\ref{fig:vot-plot}. Among all the compared trackers, our RPCF method has a 31.6\% EAO score which improves the ECO method by 3.5\%. Also, our tracker has the best performance in terms of robustness measure among all the compared trackers.
 
%------------------------------------------------------------------------
\section{Conclusion}
In this paper, we propose the ROI pooled correlation filters for visual tracking. Since the correlation filter algorithm does not extract real-world training samples, it is infeasible to perform the pooling operation for each candidate ROI region like the previous methods.
Based on the mathematical derivations, we provide an
alternative solution for the ROI-based pooling with the circularly constructed virtual samples. Then, we propose a correlation filter formula with equality constraints, and develop an efficient ADMM solver in the Fourier domain.
Finally, we evaluate the proposed RPCF tracker on OTB-2013, OTB-2015 and VOT-2017 benchmark datasets. Extensive experiments demonstrate that our method
performs favourably against the state-of-the-art algorithms on all the three datasets.

\textbf{Acknowledgement.} This paper is supported in part by National Natural Science Foundation of China \#61725202, \#61829102, \#61872056 and \#61751212, and in part by the Fundamental Research Funds for the Central Universities under Grant \#DUT18JC30. This work is also sponsored by CCF-Tencent Open Research Fund. 
{\small
\bibliographystyle{ieee}
\bibliography{egbib.bib}
}

\end{document}